\documentclass{article}

\usepackage{PRIMEarxiv}
\usepackage[utf8]{inputenc} % allow utf-8 input
\usepackage[T1]{fontenc}    % use 8-bit T1 fonts
\usepackage{hyperref}       % hyperlinks
\usepackage{url}            % simple URL typesetting
\usepackage{booktabs}       % professional-quality tables
\usepackage{amsfonts}       % blackboard math symbols
\usepackage{nicefrac}       % compact symbols for 1/2, etc.
\usepackage{microtype}      % microtypography
\usepackage{cite}
\usepackage{titlesec}
\usepackage{fancyhdr}       % header
\usepackage{graphicx}       % graphics
\graphicspath{{media/}}     % organize your images and other figures under media/ folder
\usepackage{caption}
\usepackage{subcaption}
\usepackage{amsmath}   % For mathematical symbols
\usepackage{float}     % For positioning figures
\usepackage{algorithm}
\usepackage{algorithmic}
\usepackage{indentfirst}
\usepackage{multirow}
\usepackage{eqlist}
\usepackage{enumitem}

\newcounter{alphsection}
\newcommand{\alphsection}[1]{%
  \refstepcounter{alphsection}%
  \subsection*{\alph{alphsection}.~#1}%
  \addcontentsline{toc}{subsection}{\alph{alphsection}.~#1}%
}

\DeclareMathOperator*{\argmin}{arg\,min}

\title{Constrained Transformer-Based Porous Media Generation to Spatial Distribution of Rock Properties
}

\author{
  Zihan Ren \\
  Department of Energy and Mineral Engineering \\
  Pennsylvania State University \\
  State College\\
  \texttt{zihanren.ds@gmail.com} \\
   \And
  Sanjay Srinivasan \\
  Department of Energy and Mineral Engineering \\
  Pennsylvania State University \\
  State College\\
  \texttt{szs27@psu.edu} \\
  \AND
  Dustin Crandall \\
  National Energy Technology Laboratory \\
  Morgantown \\
  \texttt{dustin.crandall@netl.doe.gov} \\
}

\begin{document}
\maketitle

%%%%%%%%%%%%%%%%%%%%%%%% abstract %%%%%%%%%%%%%%%%%%%%%%%%%%%%%%%%%%
\begin{abstract}

% importance of ct-scanning and simulation
% lack of synthetic data
% ok -> deep learning solution and conditional generation, and results
Pore-scale modeling of rock images based on information in 3D micro-computed tomography ($\mu$CT) data is crucial for studying complex subsurface processes such as CO$_2$ and brine multiphase flow during Geologic Carbon Storage (GCS). Deep learning-based approaches can successfully generate large volumes of high-resolution 3D rock microstructures such as observed in 3D micro-CT scans while honoring static rock properties. However, current state-of-the-art deep learning generative models for synthetic porous media fail to incorporate the spatial distribution of rock properties during the generation process that can have an important influence on the flow and transport characteristics of the rock. Moreover, the microstructures generated using current machine learning based approaches typically represent flow properties such as permeability or relative permeability at a scale typically below the representative elementary volume (REV) scale for those properties. In order to represent those properties at a larger scale, the spatial distribution of microstructures has to be carefully considered. Representing the spatial distribution of microstructures' properties is also critical for conditioning the models to large-scale data such as those recorded at well or field scale, which is necessary for building a consistent workflow between pore-scale analysis and field-scale modeling.

To address these challenges, we propose a two-stage modeling framework that combines a Vector Quantized Variational Autoencoder (VQVAE) and a transformer model for spatial upscaling and arbitrary-size 3D porous media reconstruction in an autoregressive manner. Our approach first compresses and quantizes sub-volume training image patches into low-dimensional image tokens using VQVAE and then trains a transformer to spatially assemble these sub-volume quantized tokens into a larger image following a certain spatial order. By employing a multi-token generation strategy, each image sub-volume is reconstructed with integrity leveraging on the transformer capabilities, and the autoregression relationship between sub-volumes is well preserved. We demonstrate the effectiveness of our multi-token transformer generation approach and validate it using real data from a test well, showcasing its potential to generate models for the porous media at the well scale using only a spatial porosity model. Compared with the reference porous medium which are unconstrained by spatial rock properties, the interpolated representative porous media that reflect field-scale geological properties give more accurate modeling of transport properties, including permeability and multiphase flow relative permeability of CO$_2$ and brine,  This approach helps bridge the gap between pore-scale models and larger geological models and improves the understanding of multiphase flow in the subsurface.
\end{abstract}

% keywords can be removed

%%%%%%%%%%%%%%%%% introduction %%%%%%%%%%%%%%
\section{Introduction} 

\label{sec:intro}

% review of the necessity of generating synthetic medium
% synthetic porous medium approach generation
% current limitations
% my approach

Developing an efficient and accurate method for generating 3D micro-CT porous media is crucial for numerous engineering applications, particularly for developing controllable digital twins that can be used to analyze their varied material properties thoroughly \cite{2020IBMresearch, 2021Niu_digitalpore} under prescribed flow and transport conditions. For instance, in Geologic carbon storage (GCS), understanding flow dynamics of CO$_2$ and brine phases is vital for modeling the displacement of the CO$_2$ plume in the subsurface and assessing long-term subsurface storage potential of the reservoir \cite{2014Burnside_krccus}. Data acquisition to determine physical properties such as relative permeability is typically done through laboratory experiments or physical simulations on 3D CT images. Although laboratory measurements are generally accurate, they are labor-intensive and time-consuming \cite{2020zhao_krpredML}. More critically, they are only representative of the property at a small scale and to the specific rock type sampled in the subsurface.\par

Performing direct pore-scale simulations on 3D micro-computed tomography (CT) images offers a more flexible approach to studying multiphase behavior at the micron scale. That approach also allows for manipulating different physical factors, such as the capillary number and viscosity, to study the sensitivity to various physical factors \cite{1988_LenormandKr,2021Prakash_krEoS}. This in turn can yield more robust calculation of multiphase flow process properties like relative permeability using pore-scale simulation such as pore network model coupled with Stokes flow simulation \cite{2004Blunt_PNM}. However, micro-CT samples are expensive, sparse, and representative of a small portion of the variations in subsurface geologic properties. Simulation results on micromodels (miniaturized artificial pore network model) \cite{2016Senyou_psonkrk, 2014Xu_poregeoonkr} as well as laboratory experiment results \cite{2013Zhang_krheterogeneity} reveal that pore structure properties such as pore throat ratio, coordination number, shape factors, and pore throat orientation, as well as geological heterogeneity such as spatial porosity distribution, all have a significant impact on relative and absolute permeability. The pore-scale models constructed on the basis of micro-CT scans, however, lack a direct connection to larger geological models that incorporate field-scale geological properties variations. Therefore, building a representative pore-scale model that can capture both pore-scale physics but can be assembled to reflect large-scale geological property variations found at well or even field scale is necessary for linking the pore scale observations to the field scale.

% listed traditional approach and deep learning simple models
To address some of the above issues, research efforts have focused on developing synthetic models for porous media to alleviate the cost of acquiring micro-CT scans and to increase the variability and abundance of pore-scale models. Statistics-based techniques such as those based on multipoint statistics are able to generate models that preserve the continuity of features such as pore throats \cite{2005_pnmreconstruct_blunt, 2010Sanjay_multipoint_reservoir, 2013Sahimi_2pointreconstruct}. However, the disadvantage of these approaches is that they are computationally expensive and are limited to the reconstruction of complex and heterogeneous object patterns under the conditions of statistical stationary and the availability of rich training data \cite{2017MosserGAN}. Recent advancements in deep learning offer potentially more comprehensive solutions to reconstruct complex target distributions while assimilating different data sources by using deep generative neural networks. Some successful examples include using AttentionGAN in image synthesis \cite{2017TaoXu_AttnGAN}, 2D/3D image rendering \cite{2015_vaeDCIGETejas}, scene reconstruction \cite{2020Nerf}, and text-to-image/video models driven by latent diffusion transformers \cite{2023Gao_PreDiff, 2023shiyunyu_videolatent}. These models have primarily focused on building generative models conditioned on semantic meanings, especially prompts, tokens, and spatial structure information. Compared with stochastic geostatistical reconstruction methods such as those based on multiple-point statistics (MPS) \cite{2007Sanjay_multipoint}, deep learning-based generative models are more capable of capturing complex patterns from training data \cite{2017MosserGAN} since they have a larger parameter space.\par

% listed applications in subsurface
In the realm of subsurface properties reconstruction, deep learning approaches are widely used in channel reconstruction \cite{2020ChanGANsubchannel}, solving inverse problems \cite{2022wu_vaeinverse, 2022_gan_RL}, and conditional reservoir model generation \cite{2020SongGANSIM, 2022Song_GANSim3D}. In the realm of digital rocks, GAN-based deep learning models such as Deep Convolutional Generative Adversarial Networks (DCGAN) have been utilized to reconstruct 2D and 3D micro-CT images of Berea sandstone, bead packs, and oolitic Ketton limestone \cite{2017MosserGAN, 2018MosserGAN, 2021AssembleGAN_Sung} while honoring petrophysical and Minkowski functional properties from training samples. Recently, some novel frameworks such as DiAGAN \cite{2020Guillaume_DiAGAN}, SliceGAN \cite{2021GAN_2D3D_stevekench}, or denoising diffusion-based techniques \cite{2024phan_2d3d_denoising} have been used to generate 3D volumes using only 2D data sections. In these approaches, a 2D convolutional discriminator is trained along with a 3D generator. However, building representative porous media that reflect certain geological properties requires more granular control over stochastic processes in GAN. A GAN and actor-critic reinforcement learning framework~(GAN-AC) was developed to search for stochastic parameters to control user-defined GAN generation \cite{2022_gan_RL}. Meanwhile, some other research focuses on deforming GAN latent space to constrain GAN generation to physical simulation outputs \cite{2016Alec_DCGANlatentspace, 2024Ren_deformGAN}.

% criticize the above approach's limitations
These deep learning models achieve reconstruction authenticity while constraining the neural network generation process to reproduce target rock properties such as porosity or permeability. However, there are two main limitations to the above methods. First, most existing research results focus on constraining 3D structures to static rock properties such as porosity, whereas most field-scale geological models are rarely homogeneous. Multiphase flow functions like relative permeability shall be linked to spatial perturbation of rock properties to perform more accurate upscaling of flow functions. Secondly, the scale of these reconstruction models tend to be restricted, mostly ranging from $64^3$ to $128^3$ voxels due to computational limits. With better computational resources, a $256^3$ voxel image can sometimes be achieved. However, as size increases, the computational resources required for data processing in terms of the number of GPUs, and their memory requirements increase, while at the same time, the number of training images becomes smaller. Most current deep learning-based generators for pore-scale models are trained on sub-volumes of a complete micro-CT data set, and larger training images result in fewer sub-volumes for model training. More importantly, when the models are synthesized at the scale of the sub-volume, most physical attributes simulated from these porous media are below the Representative Elementary Volume (REV) scale. This is especially true for transport properties such as relative permeability and permeability \cite{2020Jackson_krREV}, whose REV scale is larger than that of porosity. Analyzing transport properties below REV scale is easily influenced by local pore structure, rendering them not representative of field-scale quantities. A possible solution to the above limitation is to utilize a VQVAE-based model to create larger rock images \cite{2022Phan_2d3d_vqvae}. However, although this approach can yield large 3D CT images, the reconstruction is predicated on the availability of 2D images while honoring vertical porosity distribution instead of a more comprehensive 3D porosity model. Each individual sub-volume is sampled independently and to impose spatial continuity, a convolutional merging strategy is adopted to add each independent sampled sub-volume. While this may solve the problem of boundary transition, the VQVAE itself doesn't learn effective spatial dependencies (especially in terms of 3D positions) between different sub-volumes, thus the accuracy of reproduced features in the reconstructed models cannot be ensured in those approaches. Therefore, an improved workflow is needed for generating representative porous media conditioned to 3D spatial variations of geologic properties like porosity. These spatial variations may be reflective of the variations at the field scale and representing those variations while ensuring interconnectivity between generation blocks can yield flow properties more representative of field scale quantities.

To address some of the above challenges, we propose combining a generative-based approach with an autoregressive-based model to develop a framework for reconstructing 3D porous media with arbitrary size and flexibility to perform spatial upscaling and downscaling. In this workflow, the autoregressive-based model can dynamically generate porous media much larger than the scale of the original training image with spatial continuity. We start by compressing the smaller-size training images into low-dimensional tokens using the VQVAE/VQGAN framework \cite{2017Aaron_VQVAE, 2020Patrick_VQGAN}. In the VQGAN framework \cite{2020Patrick_VQGAN}, a transformer autoregressively generates each image token quantized by the VQVAE model. However, in our framework, instead of focusing on a single image, we subdivide a larger image into smaller sub-volumes matching the size of the VQVAE training image. Subsequently, we quantize each image patch into tokens and train a transformer to spatially assemble those low-dimensional codebook tokens constrained by spatial rock properties. Inspired by multi-token transformer generation strategy \cite{2024Fabian_multitoken_transformer}, our transformer model generates multiple tokens per image patch, preserving image integrity and allowing the transformer to focus on figuring out the autoregression relationship between patches.\par

The contributions of our work can be summarized as follows:
\begin{itemize}[noitemsep, topsep=0pt, partopsep=0pt, parsep=0pt]
    \item Develop and demonstrate a method based on VQVAE and transformer two-stage training framework for generating 3D porous medium that reflects the spatial variations in a 3D porosity model using autoregression.
    \item We demonstrate that the transformer model can be very effective at the multi-token generation task, especially for tokens that embody complete sub-image patch features.
    \item Our approach has been validated using real data from a well, showcasing its potential to interpolate CT scans at the well scale using only a spatial porosity model.
    \item Our proposed workflow can accurately interpolate and calculate representative permeability and relative permeability values by incorporating spatial heterogeneity of geological properties such as porosity.
\end{itemize}

\begin{figure}[hbt]
    \centering
    \includegraphics[width = \textwidth]{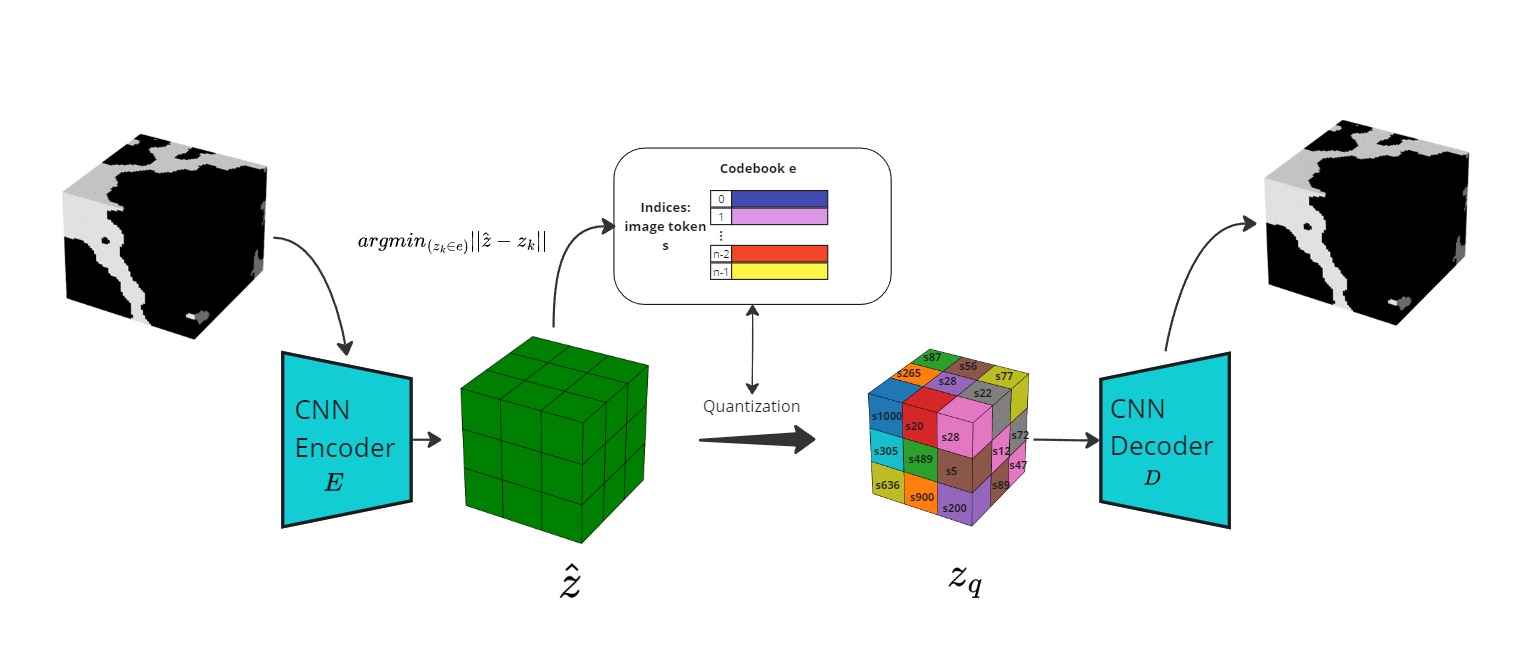}
    \caption{The workflow of Vector Quantized Variational Autoencoder
}
    \label{method_fig:VQVAE_workflow}
\end{figure}\par

%%%%%%%%%%%%%%%%%%%%%%% Method %%%%%%%%%%%%%%%%%%%%%%%%%%%%%%

\section{Methods}
\label{sec:Method}

\subsection{Vector Quantized Variational Autoencoders}
\label{subsec:method/VQVAE}

In order to have the ability to generate models for the porous media at any arbitrary scale, an autoregressive modeling framework is needed. Some well-known image autoregression models such as PixelCNN \cite{2016_pixelCNN} can generate pixel-by-pixel renditions in sequential fashion. However, PixelCNN operates directly at the pixel scale, making the generation of larger images challenging while maintaining global coherence \cite{2020Patrick_VQGAN}. This limitation renders it a non-ideal candidate for generating 3D models of porous media micro-structures at arbitrary scales. Building autoregressive models in the latent space rather than at the pixel scale is more computationally feasible and facilitates control over the coherency and conditioning of the resultant models, especially in the context of video generation \cite{2023Gao_PreDiff, 2023shiyunyu_videolatent}. However, constructing a process for training within the latent space in order to develop image tokens and subsequently, feeding those image tokens to autoregression models to ensure coordination between the image token compression model and the autoregression model can be challenging.\par

A two-stage approach using Vector Quantized Variational Autoencoder (VQVAE) is an ideal candidate to address these challenges \cite{2017Aaron_VQVAE, 2020Patrick_VQGAN}. It uses a vector quantized variational autoencoder to compress images to latent space and discretize latent space using codebook indices. Autoregression-based models such as transformers can directly model codebook indices instead of image tokens to build the autoregression process. In our approach, we use VQVAE to compress 3D microstructures of rock into latent vector space $\hat{z}$~(Equation~\ref{equ:encoderfunc}), which consists of a set of latent feature vectors $\hat{z} = \{ \hat{z_1}, \hat{z_2}, ..., \hat{z_t} \}$, where $t$ is the total number of feature vectors per image. The feature vectors are created by applying the 3D Convolutional Neural Network~(CNN) encoder function $E$ on the array $x$ of the 3D microstructures of voxel dimensions $l$, $w$, and $d$.
\begin{equation}\label{equ:encoderfunc}
\hat{z} = E(x) \in \mathbb{R}^{l \times w \times d}
\end{equation}
The VQVAE workflow is described in Figure~\ref{method_fig:VQVAE_workflow}. After encoding the image to latent vector space, all latent vectors inside $\hat{z}$ are discretized by mapping onto the nearest element of codebook $(e)$ entry $z_k$ as described in Equation~\ref{equ:codebook}, where we use $q(\hat{z})$ to represent the mapping process. Codebook $e$ is an embedding space $e \in \mathbb{R}^{K \times D}$, where $K$ is the size of the embedding space and $D$ is the dimensionality of each latent embedding vector, which is the same as our compressed image latent vector $\hat{z}$.
\begin{equation}\label{equ:codebook}
z_q = q(\hat{z}) = \argmin_{z_k \in e} |\hat{z}_{i} - z_k|
\end{equation}
The quantization process maps approximated latent space $\hat{z}$ to latent space $z_q = \{ z_q^1, z_q^2, ..., z_q^t \}$ sampled from codebook $e$, with the corresponding codebook indices $\mathcal{S} = \{ s_1, s_2, ..., s_t \}$. The mapped latent space should have the same number of latent vectors and dimensions as the original $\hat{z}$. The reconstruction of 3D rock microstructure will be upsampled by a 3D Convolutional Neural Network~(CNN) decoder: $\hat{x} = G(z_q) = G(q(\hat{z}))$. The loss function mainly comprises reconstruction loss and codebook regularization as described in Equation~\ref{equ:lossvqvae}, where $sg[\cdot]$ denotes the stop-gradient operator. Loss terms $\alpha|sg[E(x)] - z_q|_2^2 + \beta|sg[z_q] - E(x)|_2^2$ ensure that the encoder outputs and codebook vectors are close to each other. The latter loss term is also called commitment loss. The VQVAE discretization procedure can compress and discretize image latent space, and the compressed image tokens, which are the codebook indices $\mathcal{S}$ set, are suitable for autoregression-based models such as transformers to process, similar to natural language processing tasks.
\begin{equation}\label{equ:lossvqvae}
\mathcal{L}_{VQ} (E,G,q) = |x-\hat{x}|^2 + |sg[E(x)] - z_q|_2^2 + |sg[z_q] - E(x)|_2^2
\end{equation}

\begin{figure}[hbt]
    \centering
    \includegraphics[width = \textwidth]{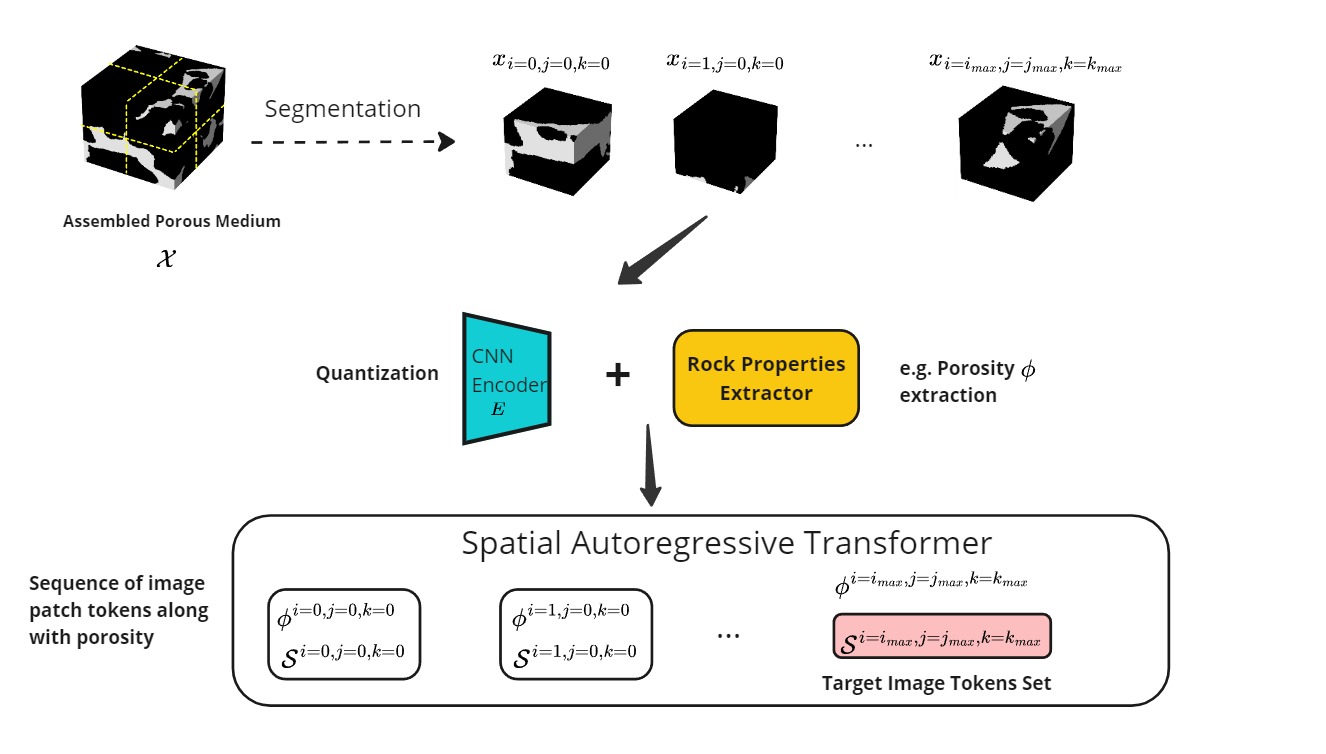}
    \caption{Workflow of the spatially assembled transformer. This process models spatial autoregression among patches $x$, associated with coordinates $(i,j,k)$ within a large porous medium $\mathcal{X}$. The transformer is trained on both the compressed spatial image token set $\mathcal{S}$ and corresponding rock properties, in this case porosity $\phi$. The transformer predicts the last patch token set $\mathcal{S}$ given all current and previous porosity information and all previous token sets $\mathcal{S}^{i=0,j=0,k=0}, ..., \mathcal{S}^{i=n,j=n,k=n-1}$, where n is the dimension of $\mathcal{X}$. The encoder $E$ was pretrained during the VQVAE workflow. Rock properties can be extracted using either an image processing tool to extract static properties like porosity or a pore network model to obtain flow-related variables. For simplification, we use the image processing tool PoreSpy \cite{2019porespy} to extract porosity.
}
    \label{method_fig:transformer_workflow}
\end{figure}\par

\subsection{Transformer}
\label{method:transformer}
Transformer-based models are excellent at handling Natural Language Processing~(NLP) related tasks such as translation or language modeling, which require processing or generating sequences of language tokens. A key advantage brought by transformer-based models for processing sequential data is their self-attention mechanism, which is adept at both parallel processing of different positions in a sequence and handling long-range dependencies \cite{2017Transformer}. Self-attention in transformers can efficiently determine which positions to focus on when predicting current outputs. The self-attention layer primarily consists of three matrices: query~($Q$), key~($K$), and value~($V$). The attention features are calculated through a scaled dot product process as described in Equation~\ref{equ:transformer}, where $\sqrt{d_k}$ is the scaling factor and $softmax(\frac{QK^T}{\sqrt{d_k}})$ calculates the attention score or weight that can be assigned to the value corresponding to every position combination.\par
\begin{equation}\label{equ:transformer}
Attention(Q,K,V) = softmax(\frac{QK^T}{\sqrt{d_k}})V
\end{equation}
Self-attention computation can be repeated several times to form a multi-head attention layer, which simply concatenates different self-attention features to produce final outputs. This allows the transformer to attend to different positions under different contexts. This mechanism is similar to how different kernels in CNNs can extract various image feature maps. In addition to the multi-head attention layer, a fully connected position-wise feedforward neural network uses linear transformations to process each position individually.\par

There are two main types of transformer models in terms of application: autoregressive-based transformers (decoder-only) like the Generative Pretrained Transformer~(GPT) series \cite{2019GPT_2} and bidirectional encoder representation from transformers like BERT \cite{2019BERT}. Autoregressive-based transformers attempt to predict data in future positions given all previously available sequential features. Famous applications such as GPTs are trained to predict the last token of input language data. Due to their autoregressive nature, this transformer architecture masks the attention matrix to the lower diagonal to prevent future information from influencing current state predictions. In contrast, encoder-based models like BERT are more suitable for processing entire sequences, such as language translation or sentiment analysis of a sentence\cite{2019BERT}. While transformers are primarily used as processing language tokens for NLP tasks, they are also widely used in image processing and generation \cite{2021Alex_VIT}. However, a key drawback of using transformers for image reconstruction is that computational costs increase quadratically with sequence length \cite{2020Patrick_VQGAN}. Pixel-based image data typically have much longer sequences than a typical English sentence, making transformers less favorable for modeling images.\par

\subsection{Learning Spatial Assembled Multi-Token Transformer}
\label{subsec:method/spatial_transformer}
The two-stage approach, combining VQVAE and transformer, addresses some of the computational challenges associated with image reconstruction by training the transformer on compressed image tokens (codebook indices $s$) rather than voxel space. The approach involves using the transformer to autoregressively predict a sequence of codebook indices $p(s_1, ..., s_t)$ that have been quantized by the VQVAE model for a single image. This requires maximizing the log-likelihood of the generated image tokens, as described in Equation~\ref{equ:transformer_loss}, where $t$ represents the total number of tokens in the sequence. That amounts to minimizing the cross-entropy loss across all tokens in the sequence as given by the objective function in Equation~\ref{equ:transformer_loss}.
\begin{equation}\label{equ:transformer_loss}
\mathcal{L}_{\text{Transformer}} = \mathbb{E}_{x\sim p(x)}\left[-\sum_{i=1}^t \log p(s_i|s_{<i})\right]
\end{equation}
In the original VQVAE-Transformer training framework, the transformer is typically trained to predict the last token, which corresponds to only one feature vector $\hat{z_i}$. The overall framework can effectively assemble latent features (total sequence features length is $t$) for completely representing the 3D porous media. In our experiment, we set the number of feature vectors as 64 ($t=64$). However, as mentioned in section~\ref{sec:intro}, our goal is to reconstruct the spatial 3D microstructure of porous medium where a property such as porosity changes throughout the volume of the media at an arbitrary scale. Reconstructing a single image token sequence is not sufficient to perform systematic spatial upscaling.

Recently, multi-token transformers have gained popularity due to their better performance and faster training speed compared with single-token transformers \cite{2024Fabian_multitoken_transformer}. Instead of generating an individual image token $s_i$, a multi-token transformer can generate multiple image tokens per inference, which is defined as $\mathcal{S}$ of this set, where $\mathcal{S} = \{ s_1,s_2, ..., s_t \}$. Similarly, we designed a spatially assembled multi-token transformer that aims to learn the autoregression relationship among different individual subvolumes of 3D microstructure porous medium. We define a large 3D porous medium as $\mathcal{X}$ which consists of individual subvolumes $x_i \in \{ x_1, x_2, ..., x_n \}$. In this scenario, the goal of the transformer is to autoregressively generate a sequence of tokens set $ { \mathcal{S}_1, \mathcal{S}_2, ..., \mathcal{S}_n } $ and modeling spatial dependencies among the latent space representing image patches. Since we have a total of $n$ image patches per large image $\mathcal{X}$ and each image can be quantized into $t$ image tokens, the total sequence length will be $t \times n$. Thus, we are trying to predict $ p(s_{(i+1)*t:i*t}|s_{j<i*t}) $, where $t$ is the length of the image tokens sequence for one individual image patch and $i$ represents the $i$th image patch along $\mathcal{X}$. To simplify notation, we use $\mathcal{S}$ as a replacement for the set of patch-based image tokens, then the target becomes $p(\mathcal{S}_i | \mathcal{S}_{j<i})$, which we can define as our current transformer loss function in Equation~\ref{equ:transformer_loss_multitoken}.

\begin{equation}\label{equ:transformer_loss_multitoken}
\mathcal{L}_{\text{Transformer}} = \mathbb{E}_{\mathcal{X}\sim p(\mathcal{X})}\left[-\sum_{i=1}^n \log p(\mathcal{S}_i|\mathcal{S}_{j<i})\right], \mathcal{S}_i = { s_1, ..., s_t  }
\end{equation}

\subsection{Constraining generation to spatial rock properties}
Constraining 3D porous media microstructure generation to spatial rock properties is crucial to represent flow functions at larger scales. We can simply append the conditioning rock properties $\mathcal{C}_i$ along with image tokens during the training process. This is represented in Equation~\ref{equ:transformer_loss_multitoken_cond}.
\begin{equation}\label{equ:transformer_loss_multitoken_cond}
\mathcal{L}_{\text{Transformer}} = \mathbb{E}_{\mathcal{X}\sim p(\mathcal{X})}\left[-\sum_{i=1}^n \log p(\mathcal{S}_i|\mathcal{S}_{<i},\mathcal{C}_{\leq i})\right], \mathcal{S}_i = { s_1, ..., s_t  }
\end{equation}

Each $\mathcal{C}_i$ is the constant conditioning information at a certain grid block expanded to match with single subvolume token length $t$. The process for constraining transformer generation to spatial rock properties can be described in Figure~\ref{method_fig:transformer_workflow}, where we use porosity $\phi$ as an example of the spatial conditioning data $\mathcal{C}_i$. Any other field scale data that relates to the characteristics of the porous media can be used as conditional information. Each segmented subvolume $x$ from $\mathcal{X}$ has associated grid coordinates $(i,j,k)$ and conditioning information $\mathcal{C}_i$ will be appended to the image token sequence. For each image subvolume $x_i$, we use the pretrained VQVAE encoder to get the image tokens set $\mathcal{S}$. Along with conditioning data, we can get a long sequence of image tokens consisting of different $\mathcal{S}$ sets. The goal for the transformer is to predict the last image token set $\mathcal{S}^{i=i_{max},j=j_{max},k=k_{max}}$, given previous token sets $
\{ \mathcal{S}_1^{i=0,j=0,k=0}, ...,  \mathcal{S}_{n-1}^{i=i_{max},j=j_{max},k=k_{max}-1} \}
$ as well as all spatial conditional information (including current state conditional information) $
\{ \mathcal{C}_1^{i=0,j=0,k=0}, ...,  \mathcal{C}_n^{i=i_{max},j=j_{max},k=k_{max}} \}
$. Both input sequence vector and conditional sequence vector have the same sequence length $t \times n$, as we introduced in section~\ref{subsec:method/spatial_transformer}. We append an SOS token set $\mathcal{S}_0$ to the beginning of the real $\mathcal{S}$ token set to initialize inference. The structure of the input and output of our multi-token transformer can be summarized in Equation~\ref{equ:transformer_input_output}.

\begin{equation}\label{equ:transformer_input_output}
\begin{aligned}
    \text{tokens input} &= \{\mathcal{S}_{sos}, \mathcal{S}_1, \ldots, \mathcal{S}_{n-1}\} \\
    \text{cond input} &= \{\mathcal{C}_1, \mathcal{C}_2, \ldots, \mathcal{C}_{n}\} \\
    \text{output} &= p(\mathcal{S}_n \mid \mathcal{S}_{<n}), \ \ \mathcal{S}_n = \{s_1^n, s_2^n, \ldots, s_t^n\} \ \ \ i=i_{max},j=j_{max},k=k_{max}
\end{aligned}
\end{equation}

\subsection{Spatial Sliding Attention Window}
The spatial information is incorporated into the transformer via positional embeddings applied to the sequence positions, rather than directly appending the spatial coordinates \((i, j, k)\) to the conditional vector \(\mathcal{C}_i\). The ordering of the token set \(\{\mathcal{S}_1, \mathcal{S}_2, \ldots, \mathcal{S}_{n}\}\) follows a deterministic spatial pattern based on the \((i, j, k)\) coordinates. Specifically, let \(i_{\text{max}}\), \(j_{\text{max}}\), and \(k_{\text{max}}\) denote the maximum values for each spatial dimension within the attention window. The tokens are ordered according to a nested iteration over these dimensions:

\begin{enumerate}
    \item Iterate \(i\) from \(0\) to \(i_{\text{max}}\),
    \item For each \(i\), iterate \(j\) from \(0\) to \(j_{\text{max}}\),
    \item For each \((i, j)\) pair, iterate \(k\) from \(0\) to \(k_{\text{max}}\).
\end{enumerate}

This sequence of nested loops ensures that all grid coordinates $(i,j,k)$ are systematically traversed in the specified order using a sliding attention window. During training, the transformer implicitly learns the spatial assembly of tokens following this ordering of the token set\(\{\mathcal{S}_1, \mathcal{S}_2, \ldots, \mathcal{S}_{n}\}\). However, for reconstructing a porous medium at an arbitrary scale, a single attention window is insufficient due to the sequence length limitations. This limitation can be addressed by using a dynamic spatial sliding attention window, which moves across the spatial domain in the same traversal order as the token sequence within a single attention window. As the attention window shifts to a new position, the previously generated token sets from earlier attention windows serve as spatial conditioning data, allowing the transformer to generate new tokens at the updated location. This approach effectively extends the attention mechanism across larger spatial regions, overcoming the inherent size constraints of a fixed attention window.

\subsection{Evaluating workflow for generating flow functions}
\label{subsec:method/eval_upscale}

We use porosity as conditioning information $\mathcal{C}$ to reconstruct porous media in this experiment. Porosity is not only a common attribute in field-scale reservoir models but also serves as a first-order Minkowski function for characterizing pore structure parameters \cite{2000Mecke:Minkowski}. Thus, porosity acts as a key link in building models for porous media at arbitrary scales for evaluating both permeability and relative permeability.

Our two-stage approach, comprising VQVAE and transformer, requires evaluation of both neural networks' performance. The evaluation process consists of three main parts:

\begin{enumerate}
    \item \textbf{VQVAE Performance:}
    We assess how accurately the VQVAE's quantized vectors can reconstruct the original image representation. This involves visual inspection and comparison of statistics of original and synthetic models.

    \item \textbf{Transformer Token Quality:}
    Given our use of multi-token transformer training and inference, we evaluate the quality of these ``multi-tokens''. We examine whether the transformer-sampled tokens, when decoded into 3D porous medium subvolumes, are physically plausible and preserve the physical parameter distribution of the training subvolumes.

    \item \textbf{Assembled Porous Media Evaluation:}
    This more complex evaluation focuses on the coherency of the porous media models constructed conditioned to spatial properties. We conduct flow simulations on the assembled porous medium to compare our transformer-based predictions of transport properties with real data. Additionally, we perform two-point correlation calculations on the assembled porous media, comparing results with both real and reference data.
\end{enumerate}

The following sections detail the main physical metrics and flow function calculations used in these evaluations.

\alphsection{Two Point Probability Function}
The two-point probability function $S_2(p(\boldsymbol{x}),p(\boldsymbol{x}+\boldsymbol{h}))$ is the probability that two randomly selected points $\boldsymbol{x}$ and $\boldsymbol{x}+\boldsymbol{h}$, separated by lag distance $\boldsymbol{h}$, belong to the same phase. In porous media, phases refer to solid or void space voxels, thus it can be referred to as pore-pore two-point correlation \cite{2008Torabi_twopcurve}. We compute the two-point probability function omnidirectionally on assembled porous media using the PoreSpy \cite{2019porespy} package. The two-point probability function can effectively reflect how an assembled porous medium's spatial voxel pattern covariance function compares with the original porous medium. However, while the two-point probability function provides valuable insights into the spatial structure of porous media, it has limitations in fully characterizing complex porous systems, especially in terms of representing complex spatial connectivity that may affect transport processes.

\alphsection{Absolute permeability $k_{abs}$}
The absolute permeability ($k_{abs}$) is determined using Darcy's Law, as outlined in the system of Equations given by~\ref{methodeq:kabs}, where $\boldsymbol{u}$ is the fluid velocity, $p$ is the pressure, $\mu$ is the fluid viscosity, $L$ is the length of the porous medium, $Q$ is the volumetric flow rate, $A$ is the cross-sectional area, and $\Delta P$ is the pressure drop. In pore network modeling, Stokes flow is simulated on the porous medium to obtain the aqueous phase flow rates.

\begin{equation}
\label{methodeq:kabs}
\begin{aligned}
\nabla \cdot \boldsymbol{u} &= 0 \\
-\nabla p + \mu \nabla^2 \boldsymbol{u} &= \boldsymbol{0} \\
\frac{\mu L Q}{A \Delta P} &= k_{abs}
\end{aligned}
\end{equation}

In this experiment, permeability serves not only as an evaluation metric but also as an important outcome that our transformer-based framework is expected to predict accurately while honoring spatial porosity distribution. This ability to model single-phase permeability at any arbitrary scale is crucial, especially when core analysis or empirical correlations such as the Kozeny-Carman Equation are not sufficiently representative of spatial heterogeneity observed in real rocks. The pore network simulation to compute the absolute permeability is performed using OpenPNM \cite{2016openpnm}.

\alphsection{Relative Permeability $k_{r}$}

Relative permeability is an important parameter in reservoir simulation, used to model fluid flow and transport in subsurface reservoirs at field scale. For example, in Geological Carbon Sequestration with CO$_2$ and brine two-phase systems, relative permeability can be described by Darcy's law (Equation~\ref{methodeq:krdarcy}), where $Q_{nw}$ and $Q_w$ represent the flow rates of CO$_2$ (non-wetting phase) and brine (wetting phase), respectively, $\frac{\partial P}{\partial x}$ represents pressure gradient, $\mu$ represents dynamic viscosity, $g$ is the gravitational acceleration and $\alpha$ is the angle of the flow channel relative to the horizon, and that along with the density of the phase $\rho_{nw}$ and $\rho_w$ regulates the effect of gravity along flow direction. The relative permeability $K_{rnw}$ and $K_{rw}$ are key parameters that characterize flow behaviors between these two phases in the porous medium. These dimensionless values, ranging from 0 to 1, reflect how the presence of one fluid affects the flow of the other. As the saturation of one phase increases, its relative permeability typically increases, while the other phase decreases, thus capturing the complex interplay between the fluids in the pore space. The characteristics of how these relative permeability change with saturation are affected by factors such as pore space connectivity. In pore network modeling, Stokes flow is simulated on the percolation history of the porous medium to obtain flow rates at each saturation step, which we use to calculate $k_r$ through Darcy's Law in Equation~\ref{methodeq:krdarcy}.

\begin{equation}\label{methodeq:krdarcy}
\begin{gathered}
    Q_{nw} = \frac{-KK_{rnw}A}{\mu_{nw}}\left( \frac{\partial P_{nw}}{\partial x} + \rho_{nw}g\sin\alpha\right) \\
    Q_{w} = \frac{-KK_{rw}A}{\mu_{w}}\left( \frac{\partial P_{w}}{\partial x} + \rho_{w}g\sin\alpha\right)
\end{gathered}
\end{equation}

As mentioned previously, relative permeability is influenced by a wide range of factors, including the wettability of the grain surface, flow regime factors such as capillary number, phase-related variables (most notably saturation and phase connectivity \cite{2006BehrenModCorey,2021Prakash_krEoS}), and geological properties. In terms of upscaling $k_r$, understanding how geological properties affect the scaling characteristics of relative permeability is a prerequisite. Geological properties such as porosity are key parameters linking field-scale reservoir models with pore-scale models. Therefore, similar to permeability, $k_r$ not only serves as an evaluation metric for assembled porous media but also as a key outcome that the transformer expected to predict with the help of pore network modeling while considering spatial heterogeneity of geological properties. In our experiment, we are evaluating the spatial heterogeneity of porosity, the most common parameter used to characterize both porous media and field-scale geology models. The relationship between single-phase permeability, the trajectory of relative permeability, and the spatial distribution of porosity can be conceptually expressed as:

\begin{equation}
\{k, k_r\} = f(\phi(i,j,k))
\end{equation}

where $\phi(i,j,k)$ is the spatial distribution of porosity at coordinates (i,j,k). This conceptual function illustrates that $k_{abs}$ and $k_r$ depend on the spatial arrangement of pores (represented by the porosity distribution). Other factors such as wettability that can be manipulated during pore network modeling are not our focus in this experiment. In addition, we only perform the drainage process (i.e. the displacement of the wetting phase by the non-wetting phase in the porous medium similar to the displacement of brine by CO$_2$ during geologic sequestration) for $k_r$ simulation. In our transformer-based framework, the model receives the spatial porosity distribution as input to assemble a token set, which is then decoded by the VQVAE model to reconstruct the porous medium. This reconstructed medium, with its spatially varying porosity, forms the basis for predicting permeability and relative permeability through pore network modeling.

%%%%%%%%%%%%%%%%% Result %%%%%%%%%%%%%%%%%%%%%%%
\section{Results}
\label{sec:results}

\subsection{Dataset introduction and training workflow}
\label{subsec:result/dataset}

The training and test dataset comprises computed tomography images from the One Earth Energy Well \#1 (OEE Well \#1), which was drilled to characterize the Lower Mt. Simon Sandstone and Eau Claire Formation in the Illinois Basin \cite{2023Dustin_CT, 2024Okwen_dataupdate}. This test well, part of the Illinois Storage Corridor CarbonSAFE project, aims to assess the feasibility of safely injecting and permanently sequestering CO$_2$ produced by One Earth Energy's ethanol plant into the underlying Lower Mt. Simon Sandstone.\par

% result figures and tables cluster
\begin{table}[htbp]
\centering
\caption{CT Data for Various Core Samples}
\label{tab:ct-data}
\vspace{0.1cm}
\begin{tabular}{|c|l|c|c|c|}
\hline
\textbf{Depth (ft)} & \textbf{Name} & \textbf{Voxel Resolution ($\mu$m$^3$)} & \textbf{CT Index} & \textbf{Bulk Porosity} \\
\hline
6344 & One Earth Core3 Box2 6344 & 3.7564 & 0 & 0.15 \\
\hline
6348 & One Earth Core3 Box3 6348 & 3.7564 & 1 & 0.20 \\
\hline
6356.9 & One Earth Core3 Box6 6356.9 & 3.7564 & 2 & 0.16 \\
\hline
6371.9 & One Earth Core3 Box11 6371.9 & 3.7564 & 3 & 0.19 \\
\hline
6420 & One Earth Core3 Box27 6420 & 3.7564 & 4 & 0.14 \\
\hline
6481.5 & One Earth Core4 Box18 6481.5 & 3.7564 & 5 & 0.09 \\
\hline
\end{tabular}
\vspace{0.5cm}
\end{table}

% evaluate via scatter plot
\begin{figure}[hbt]
  \centering
  \begin{minipage}{0.45\textwidth}
    \centering
    \includegraphics[width=\linewidth]{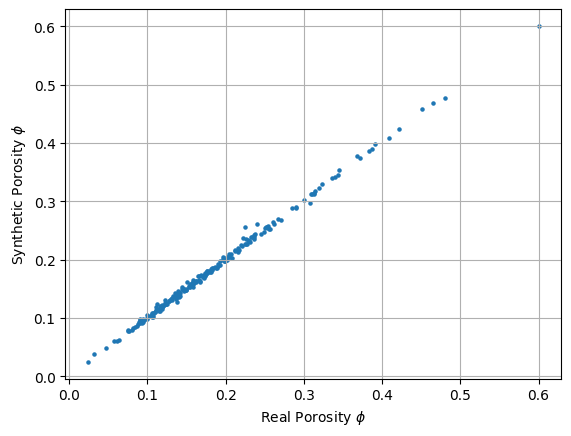}
    \caption{Scatter plot of original input porosity versus reproduced porosity in VQVAE model. The mean absolute error between the original porosity and reproduced porosity is $0.003$ }
    \label{result_fig:phivsphivqvae}
  \end{minipage}\hfill
  \begin{minipage}{0.45\textwidth}
    \centering
    \includegraphics[width=\linewidth]{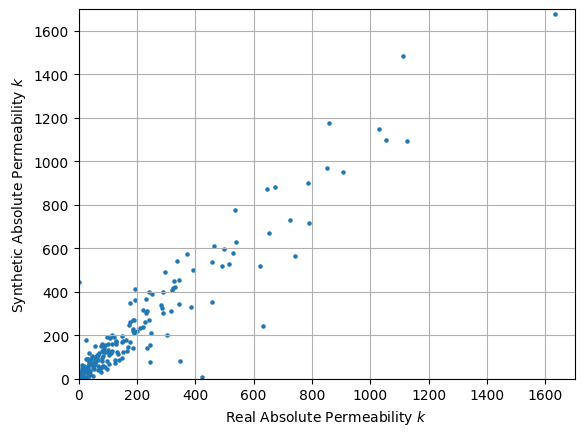}
    \caption{Scatter plot of original permeability versus reproduced permeability in VQVAE model. The mean absolute error between original permeability and reproduced permeability is $50$ md}
    \label{result_fig:kvskvqvae}
  \end{minipage}
\end{figure}

% same thing - check phi and k
% \begin{figure}[hbt]
%   \centering
%   \begin{minipage}{0.45\textwidth}
%     \centering
%     \includegraphics[width=\linewidth]{Figures/transformer_64_hist_phiphi.png}
%     \caption{Histogram of original image patch porosity~(left) versus transformer sampled image patch porosity~(right)}
%     \label{result_fig:hist_phi_vs_phi_transformer}
%   \end{minipage}\hfill
%   \begin{minipage}{0.45\textwidth}
%     \centering
%     \includegraphics[width=\linewidth]{Figures/Transformer_64_hist_kk.png}
%     \caption{Histogram of original image patch permeability~(left) versus transformer sampled image patch permeability~(right)}
%     \label{result_fig:hist_k_vs_k_transformer}
%   \end{minipage}
% \end{figure}

\begin{figure}[hbt]
  \centering
  \includegraphics[width=0.8\linewidth]{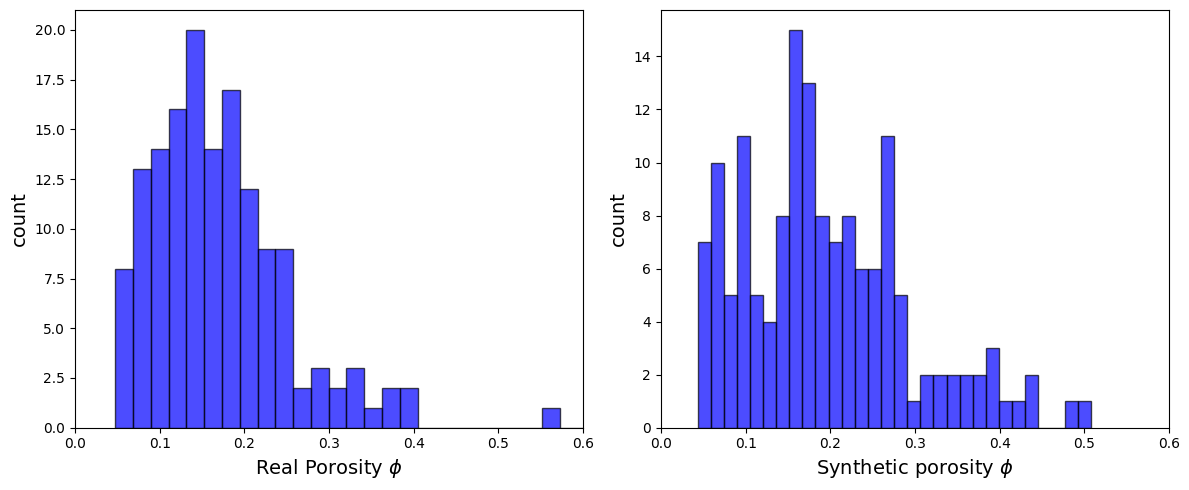}
  \caption{Histogram of original image patch porosity~(left) versus transformer sampled image patch porosity~(right)}
  \label{result_fig:hist_phi_vs_phi_transformer}
\end{figure}

\begin{figure}[hbt]
  \centering
  \includegraphics[width=0.8\linewidth]{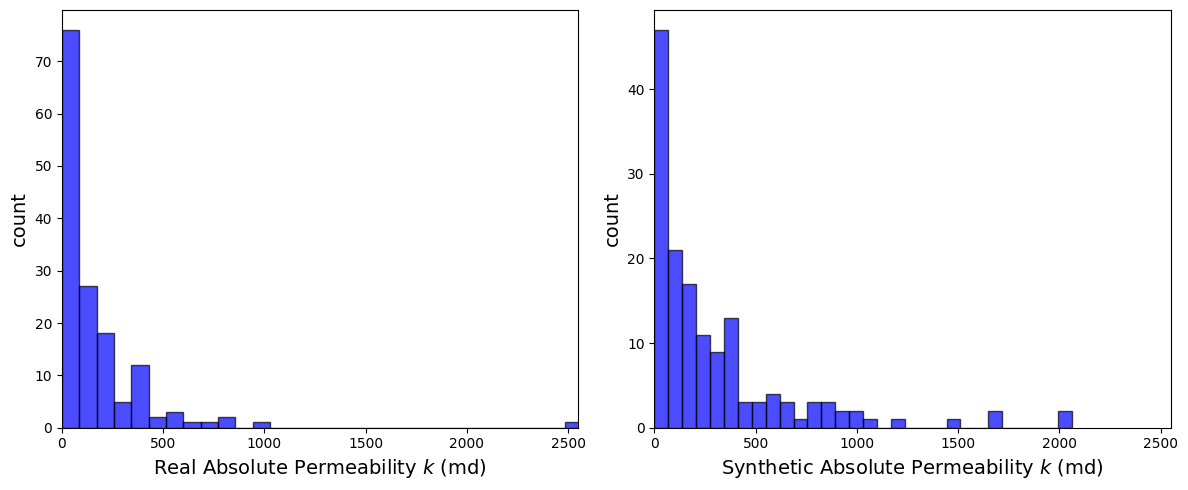}
  \caption{Histogram of original image patch permeability~(left) versus transformer sampled image patch permeability~(right)}
  \label{result_fig:hist_k_vs_k_transformer}
\end{figure}

\begin{figure}[hbt]
    \centering
    \includegraphics[width = 0.5\textwidth]{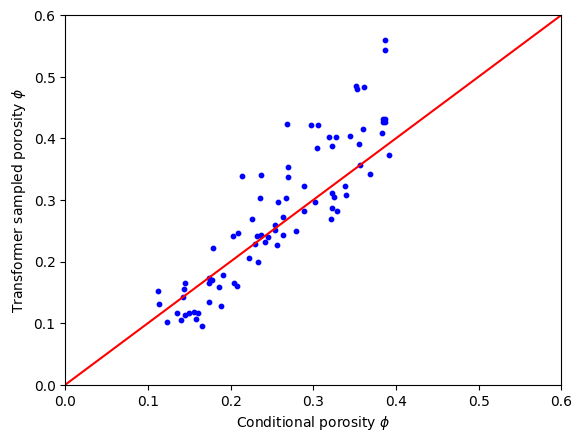}
    \caption{Conditional evaluation of transformer
}
    \label{method_fig:cond_transformer_on_phi}
\end{figure}\par

\begin{figure}[hbt]
  \centering
  \begin{minipage}{0.485\textwidth}
    \centering
    \includegraphics[width=\linewidth]{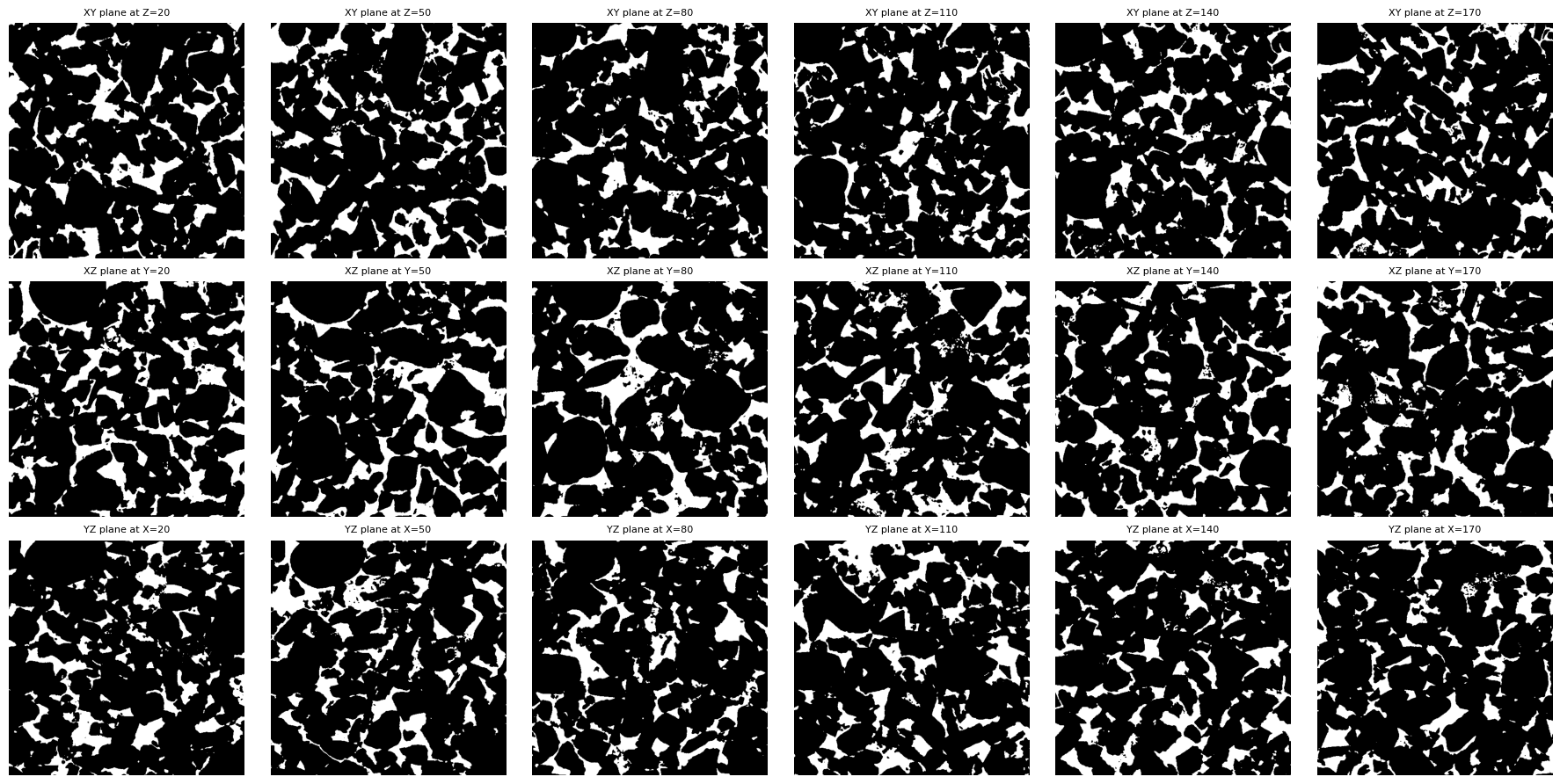}
  \end{minipage}%
  \hfill
  \vrule width 0.5pt
  \hfill
  \begin{minipage}{0.485\textwidth}
    \centering
    \includegraphics[width=\linewidth]{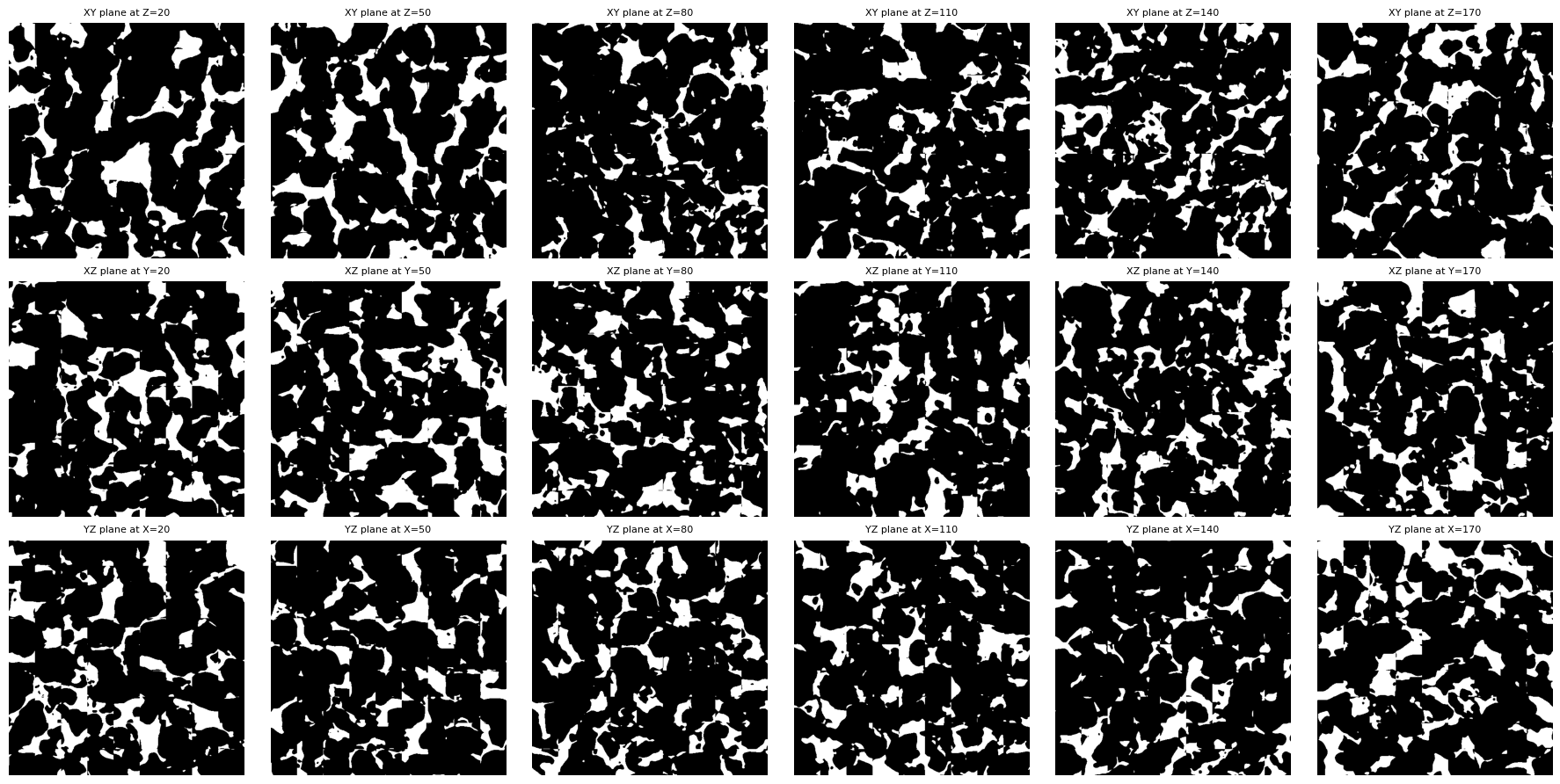}
  \end{minipage}
  \caption{Transformer assembled porous medium versus original volumes based on the same porosity spatial map. Left: original volume $576^3$ voxels; Right: same size transformer assembled porous medium.}
  \label{result_fig:transformer_assemble_fig_ctidx1}
\end{figure}

\begin{figure}[hbt]
    \centering
    \includegraphics[width = 0.7\textwidth]{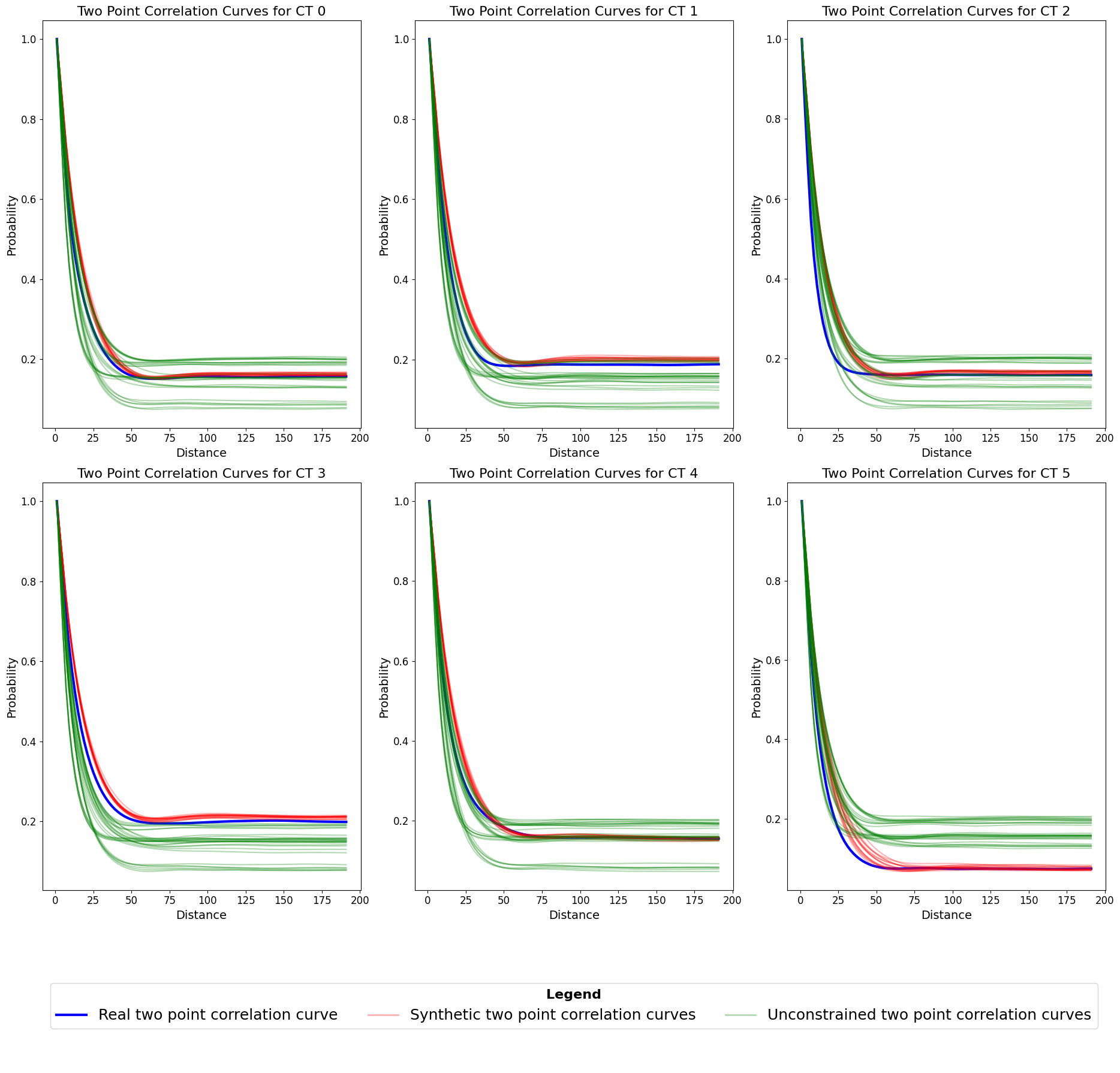}
    \caption{Two point probability curves at different CT index at volume dimension $384^3$. The green curves are two point probability curves that are not constrained by porosity, the red curves are two point probability computed curves on synthetic porous medium by transformer and the blue curve is computed on the ground truth samples
}
    \label{result_fig:two_p_curve_vol6}
\end{figure}\par

\begin{figure}[hbt]
  \centering
  \begin{minipage}{0.5\textwidth}
    \centering
    \includegraphics[width=\linewidth]{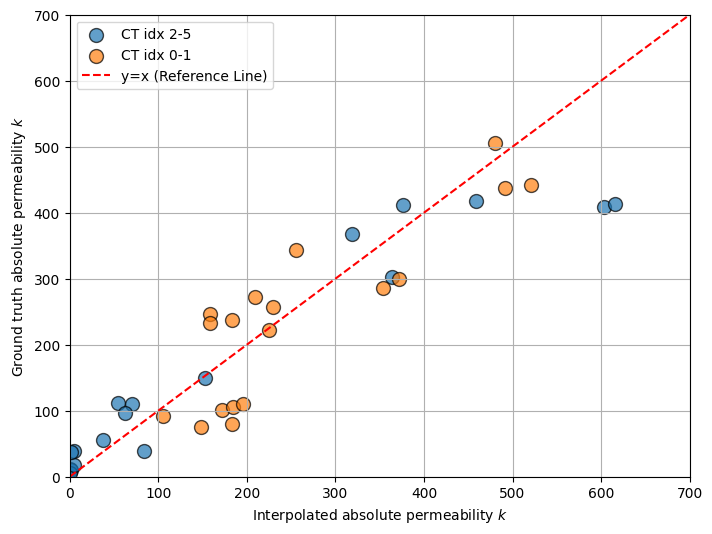}
  \end{minipage}%
  \hfill
  \begin{minipage}{0.5\textwidth}
    \centering
    \includegraphics[width=\linewidth]{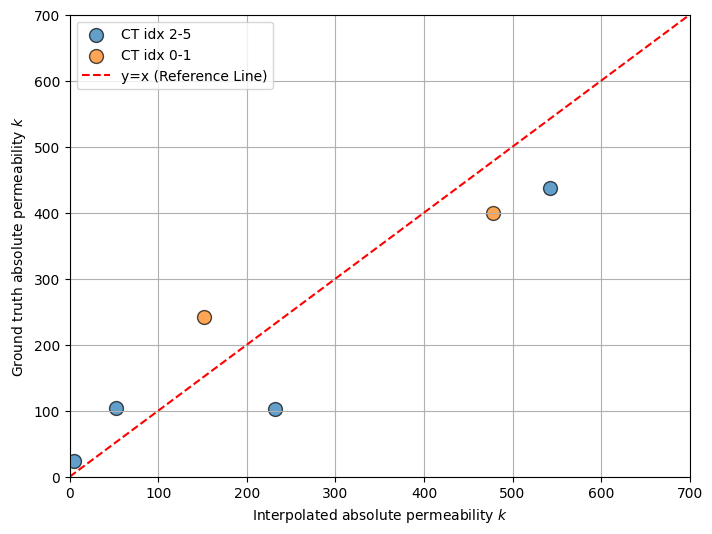}
  \end{minipage}
  \caption{Interpolated absolute permeability $k_{abs}$ versus ground truth value; Left: volume dimension at $384^3$, mean absolute error: $55$ md; Right: volume dimension at $576^3$, Mean absolute error: $65$ md.}
  \label{result_fig:interpolate_k}
\end{figure}

A total of 11 Micro-CT images were obtained using NETL's Zeiss Xradia micro-CT scanner, with resolutions ranging from 3 to 7 microns per voxel. We selected 6 3D CT volumes, all with the same voxel resolution of $3.7564 \mu m^3$, as shown in Table~\ref{tab:ct-data}. Our dataset comprises CT scans from two depth ranges: the training set spans depths from 6356.9 ft to 6481.5 ft with porosity between 0.08 and 0.2, while the test sets cover depths from 6344 ft to 6348 ft. For convenience, we rename the test sets as CT index 0 and 1, corresponding to depths of 6344 ft and 6348 ft, respectively. The training set CT scans are named as CT indices $ { 2,3,4,5 } $. 

The training process involves two key components: VQVAE and transformer. For VQVAE, we use $64^3$ subvolumes as training data~(notation $x_i$ in Figure~\ref{method_fig:transformer_workflow}), which serve as fundamental porous media components for further assembling. These sub-volumes are cropped with overlap from $635^3$ CT scans corresponding to indexed volumes 2-5, yielding a total of 32000 training samples. For the transformer, we aim to capture spatial autoregression dependencies among the smaller sub-volumes. To that end, we crop 20000 relative larger $128^3$ porous media volumes~(corresponding to $\mathcal{X}$ in Figure~\ref{method_fig:transformer_workflow}). Each $128^3$ volume is then quantized by the VQVAE's pretrained encoder into a set of 8 image tokens, along with their spatial $(i,j,k)$ coordinates in a $2^3$ grid configuration. It is important to note that each image token set $\mathcal{S}$ consists of 64 image tokens $s$, which serve as feature vectors for a $64^3$ porous media subvolume. The detailed training process and hyperparameter selection for both VQVAE and transformer are summarized in appendices~\ref{appendix:VQVAE_structure_train} and ~\ref{appendix:transformer_structure_training}, respectively. All training was conducted using an NVIDIA A6000 GPU.

In theory, our spatial assembled transformer can create an infinitely large porous medium representation. However, to make the reconstruction size comparable to the real dataset, we assembled porous media patches at two scales for evaluation: $6^3$ subvolumes ($384^3$ voxels, total 216 porous medium meta subvolumes) and $9^3$ subvolumes ($576^3$ voxels, total 729 porous medium meta subvolumes). It's worthwhile noting that a $576^3$ voxel reconstruction size is almost equivalent to an individual complete CT scan of sandstone in One Earth Energy Well \# 1. We aim to investigate whether the transformer assembler trained on CT indices $ { 2,3,4,5 } $ can create reasonable representation interpolations of CT indices 0-1 at scale $384^3$ and $576^3$, given their extracted spatial porosity distribution. But we will also evaluate the assembled reconstruction accuracy on the training CT indices $ { 2,3,4,5 } $.

\subsection{Evaluation of Quantized Vector}
The foundation of our workflow lies on the ability to construct high-quality sampled image tokens, which are subsequently used to train the spatial assembled transformer. To evaluate the quality of these image tokens, we assess the quality of quantized vectors queried by their indices. We begin by inputting segmented image patches of size $64^3$ ($x$), cropped from the test set (CT index 0 and 1), into the VQVAE encoder to obtain the approximated latent vector $\hat{z}$. This vector is then used to retrieve the most similar quantized latent vector. The quantized latent vector $z_q$ is decoded by the pretrained decoder to produce the approximated $\hat{x}$.

We perform PNM simulations on both VQVAE-decoded image patches (300) and original input image patches (300) to simulate permeability $k_{abs}$. For porosity, we use PoreSpy \cite{2019porespy} to directly calculate the void space ratio in both synthetic and original porous media. We compare the reproduced porosity $\phi$ and absolute permeability $k_{abs}$ with the original input data using scatter plots, as shown in Figures~\ref{result_fig:phivsphivqvae} and \ref{result_fig:kvskvqvae}. The porosity reconstruction proves to be highly accurate. A more scattered trend in permeability reconstruction is observed and can be attributed to the wide variability shown by the PNM simulations at this sub-REV scale ($64^3$). With only a few available percolation paths in a small pore network model, minor changes in voxel patterns can lead to dramatic changes in $k_{abs}$, as evident in some samples. However, this discrepancy is less concerning since most synthetic $k_{abs}$ values align well with the original $k_{abs}$. Assembling these porous media components into larger volumes will likely produce a more reasonable $k_{abs}$ distribution, as larger PNM simulation domains yield more representative $k_{abs}$ values. 

We also compared the synthetic image patches with the original image patches, as displayed in Figure~\ref{method_fig:VQVAE_reconstruction_compare} in appendix~\ref{appendix:image_reconstruction_simulation}. As we can see, VQVAE can successfully reconstruct most original image patch structures. The high agreement between original and synthetic reconstruction porous media patches demonstrates that our codebook has been well-trained to learn the generalized porous medium features in test set CT indices 0 and 1. A close physical similarity~($\phi$ and $k_{abs}$) between $x$ and $\hat{x}$ demonstrates the high quality of image tokens and the pre-trained codebook, indicating comprehensive and diverse porous media feature representation, particularly since the evaluation set $x$ is not part of the training set.

\subsection{Evaluation of Multi-tokens Generated by Transformer}
\label{subsec:result/transformer_single_eval}

\begin{figure}[hbt]
  \centering
  \begin{minipage}{0.5\textwidth}
    \centering
    \includegraphics[width=\linewidth]{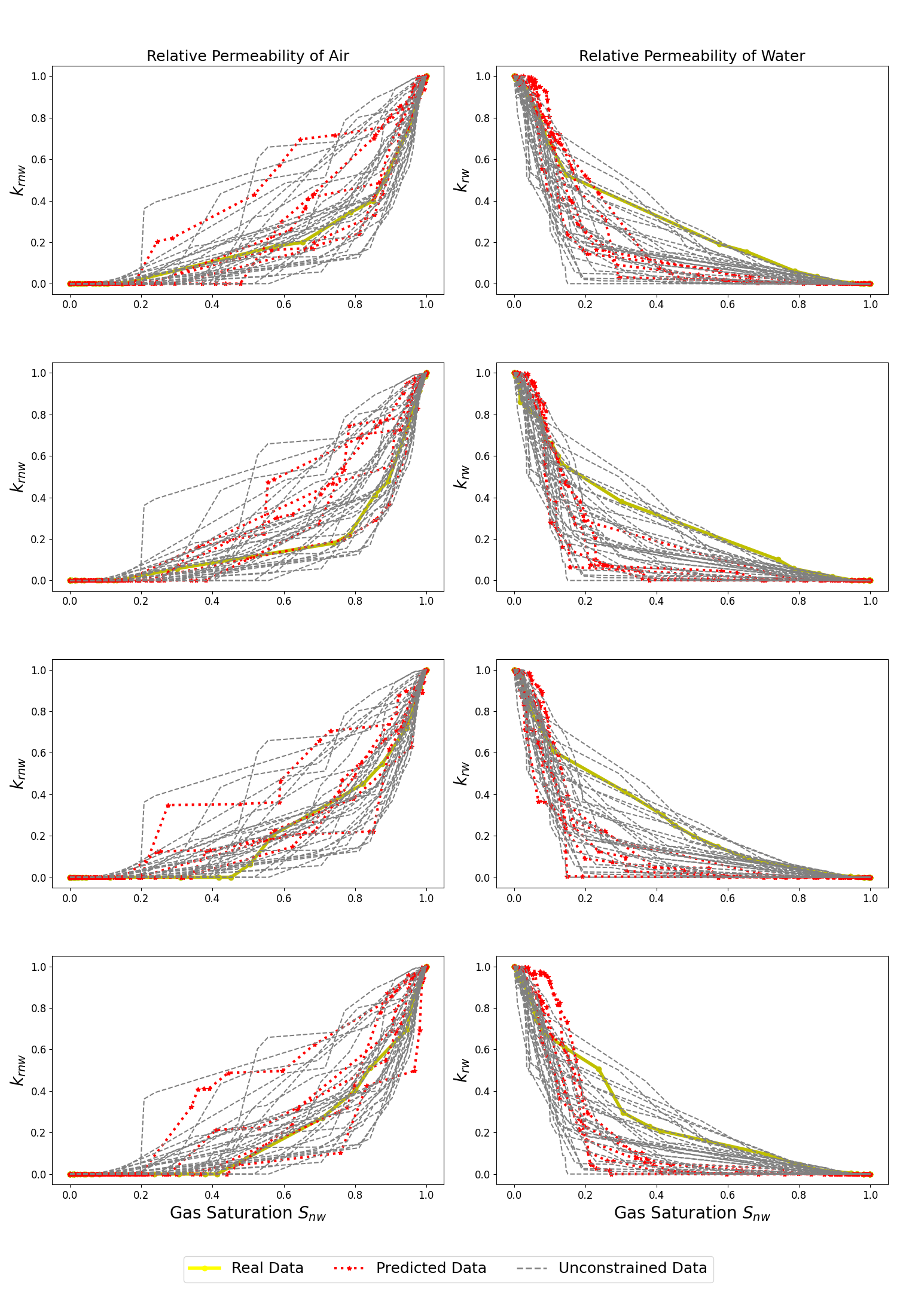}
  \end{minipage}%
  \hfill
  \begin{minipage}{0.5\textwidth}
    \centering
    \includegraphics[width=\linewidth]{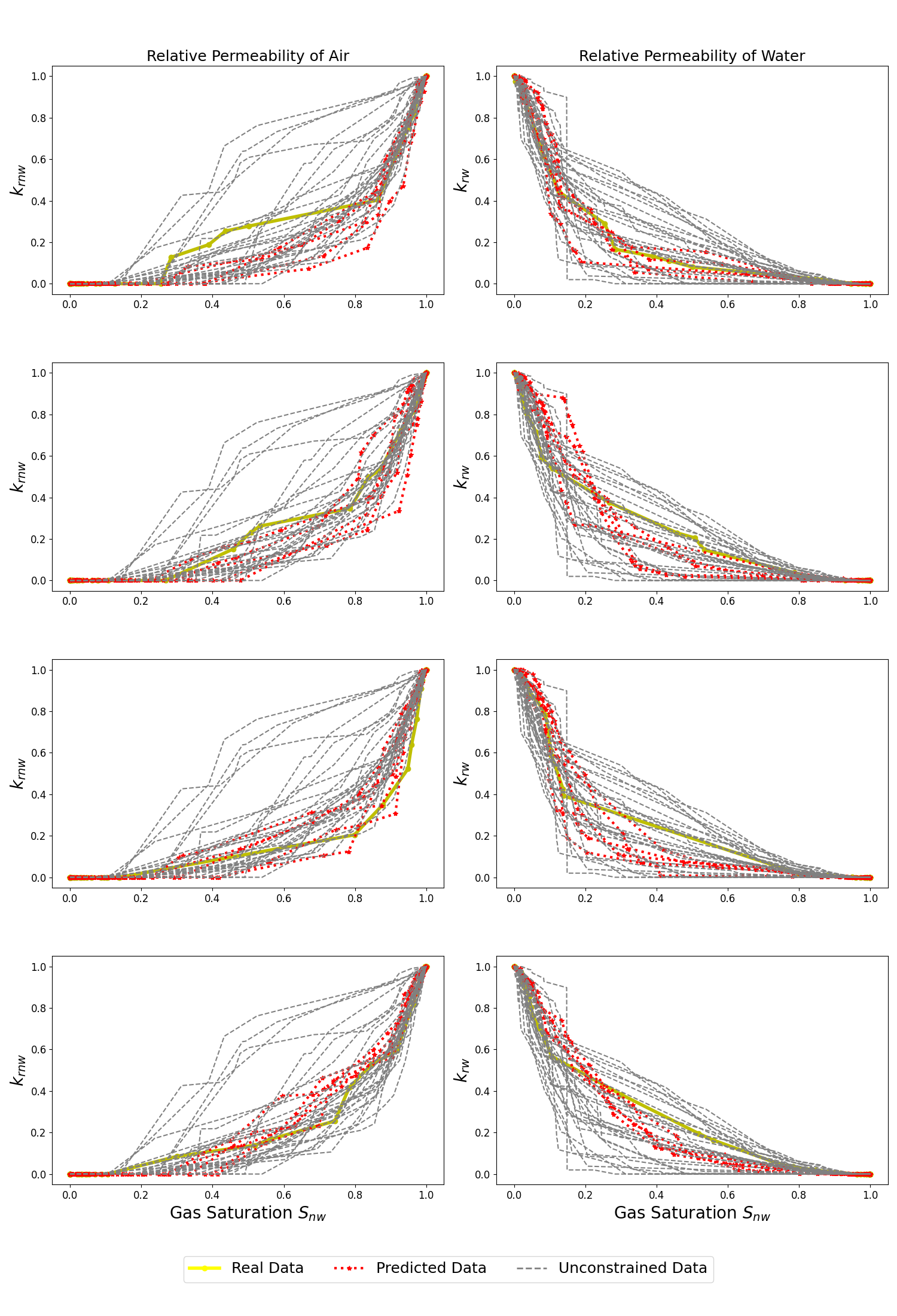}
  \end{minipage}
\caption{Relative permeability simulations on real porous medium, transformer-assembled porosity-constrained porous medium, and porous medium corresponding to other CT indices at volume dimension $384^3$ that exhibit different spatial distribution of porosity. Left: $k_r$ simulation on CT index 0; Right: $k_r$ simulation on CT index 1. Transformer-interpolated $k_r$ agrees well with ground truth curves on CT index 1 but shows poorer agreement on CT index 0}
  \label{result_fig:kr_ct_0_1_vol_6}
\end{figure}

Unlike traditional transformer single-token generation, our approach adopts a multi-token training and generation strategy. Moreover, each multi-token generation should represent meaningful features of a real porous media sub-volume of size $64^3$. In other words, we are evaluating the coherence of the volume generated by assembling patch-based token features $\mathcal{S} = \{ s_1,s_2, ..., s_t \}$. We evaluate the reconstruction distribution of our physical parameters porosity $\phi$ and permeability $k_{abs}$, as shown in Figures~\ref{result_fig:hist_phi_vs_phi_transformer} and ~\ref{result_fig:hist_k_vs_k_transformer}. Both reconstructed porosity and permeability distributions agree well with real image patch distributions, although transformer-sampled blocks tend to have slightly higher porosity compared to original statistics.

We observe similar "higher porosity" values when we assess the competency of the data conditioning process. To evaluate the efficacy of the transformer to constrain the generated porous media to specific values of porosity, we let the transformer sample series of token sets $\mathcal{S}$ conditioned on different porosity values $\phi$ and ultimately decode the tokens to obtain reconstructed porous media porosity. The comparison between decoded porosity and conditioned porosity is plotted in Figure~\ref{method_fig:cond_transformer_on_phi}. Although the mean absolute error of conditioning is 0.04, which is an acceptable range for porosity constraints, transformer-sampled images tend to generate slightly higher porosity. This may be due to the fact that during transformer multi-token generation the process is also dependent on other nearby spatial tokens set $\mathcal{S}^{i,j,k}$. During the process of assembling the porous medium, the multi-task learner tries to balance between porosity and the coherency of the reconstructed volume.

Visually, we can see that transformer-sampled images exhibit reasonable structure along different planes: XY, XZ, and YZ, as displayed in Figure~\ref{result_fig:transformer_64_img} at appendix~\ref{appendix:image_reconstruction_simulation}. Despite a slight imperfection with respect to over-estimation of porosity during the assembling process, the structure of 3D porous media is well preserved and shows a reasonable reconstruction, both from visual inspection and by checking the reproduced permeability distribution. Further evaluation and validation will be mainly discussed in section~\ref{result/eval_assemble}.

\subsection{Evaluation of Assembled Porous Medium}
\label{result/eval_assemble}

As mentioned above, transport properties simulated at a small scale ($64^3$) are noisy and easily influenced by local pore structure, as we can observe from Figure~\ref{result_fig:kvskvqvae}, which demonstrates the unpredictable behavior of $k_{abs}$ at below-REV scale. However, local pore structure influence will be minimized when evaluating the larger assembled porous media especially if the reconstruction size is comparable to the REV scale. We cropped several $384^3$ and $576^3$ target samples from CT indices 0-5. For each target sample, a spatial porosity cube was extracted at a resolution of $64^3$, which is the size of the porous medium sub-volume in our experiment. To interpolate the assembled porous medium, we simply use this spatial porosity map as conditioning information to autoregressively generate 3D larger volume simulations. Figure~\ref{result_fig:transformer_assemble_fig_ctidx1} shows transformer-based synthetic porous medium ($576^3$) versus the original sample at CT index 1. The synthetic porous medium shows good spatial continuity at a large scale without any sharp discontinuities at the boundary between different meta porous medium sub-volumes.

For each assembled porous medium, we compute a two-point probability function to evaluate their spatial continuity, perform pore network modeling, and conduct a Stokes Flow simulation to evaluate transport properties including single-phase permeability and multiphase relative permeability, which we have already introduced in section~\ref{subsec:method/eval_upscale}. Each CT index's whole volume ($635^3$) can have more diverse target sub-volumes at scale $384^3$ but a more limited set at a scale of $576^3$. At a scale of $384^3$, every CT index will have multiple samples to perform evaluation comparing with transformer-generated samples. At a scale of $576^3$, which is almost the scale of the original CT volume, each CT index only has one sample.

To validate our transformer-based reconstruction approach, we compare physical simulation results computed on different porous media samples. For each CT index under evaluation, we perform simulations on three types of data: (1) the ground truth data from the original CT scan, (2) transformer-generated synthetic porous medium constrained by the spatial porosity distribution of this specific CT index, and (3) reference data sampled from other CT indices with different spatial porosity distributions. For instance, when evaluating physical properties like $k_r$ or $S_2(x)$ on CT index 1, we first simulate these properties on the original CT index 1 data and our transformer-generated porous medium (constrained by CT index 1's spatial porosity map). We then simulate the same properties on samples from CT indices ${0,2,3,4,5}$, which corresponds to the other CT indices exhibiting different spatial porosity distribution.\par

The effectiveness of our porosity-constrained approach is demonstrated in Figure~\ref{result_fig:two_p_curve_vol6}, where the two-point probability curves from transformer-synthesized porous medium align closely with the ground truth curves. This close alignment is expected since constraining spatial porosity naturally leads to better reproduction of pore-pore covariance. The reference curves from other CT indices form an uncertainty envelope, clearly showing how porosity constraints improve reconstruction accuracy. Similar comparisons are performed for other physical properties including relative permeability, allowing us to comprehensively evaluate our model's ability to capture key physical characteristics of the porous media.\par

Figure~\ref{result_fig:two_p_curve_vol6} illustrates this concept for two-point probability curves. The transformer-synthesized porous medium, constrained by the porosity distribution, produces two-point probability functions that align closely with the ground truth curves. This alignment is expected, as constraining spatial porosity naturally leads to better reproduction of pore-pore covariance. The reference curves, derived from other CT indices, form an uncertainty envelope around these results, highlighting the improved accuracy achieved through porosity constraints. This comparison not only validates our model's performance through two-point correlation functions but also underscores the importance of incorporating spatial porosity information in porous media reconstruction.\par

\begin{figure}[htbp]
  \centering
  \begin{minipage}{0.5\textwidth}
    \centering
    \includegraphics[width=\linewidth]{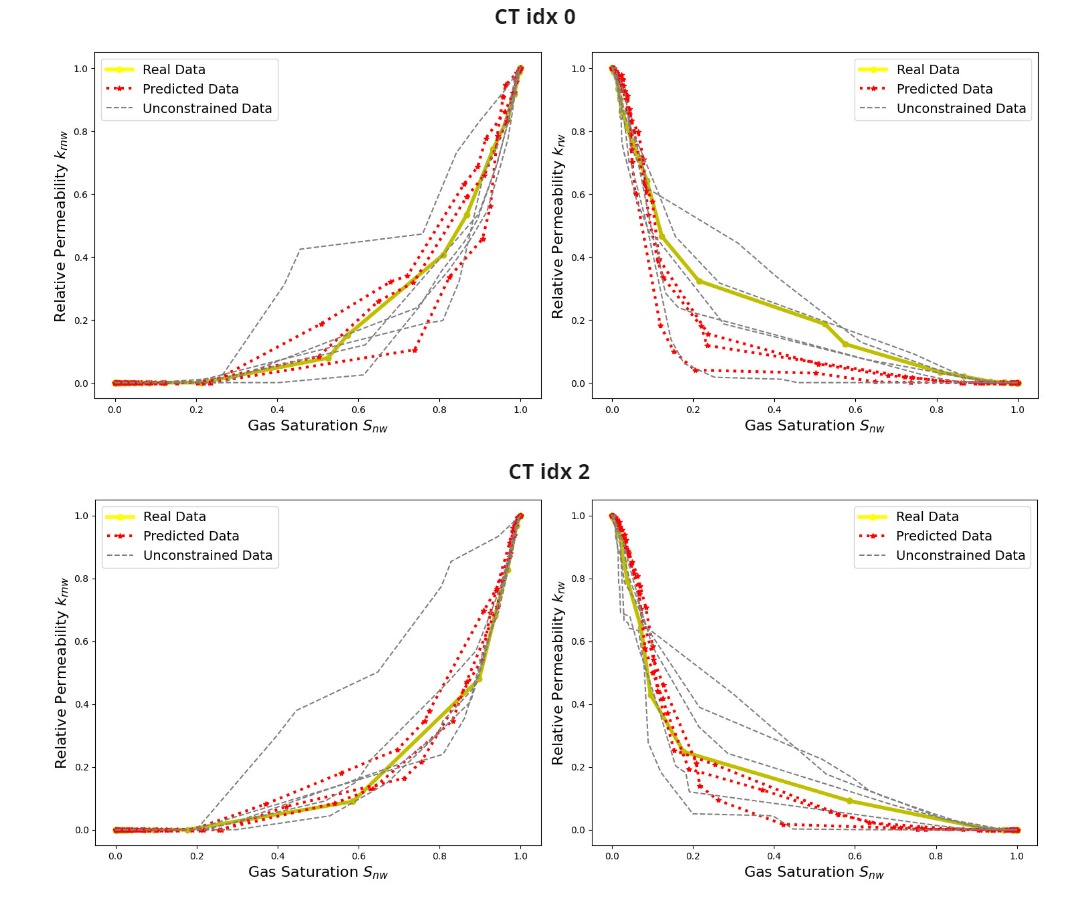}
  \end{minipage}%
  \hfill
  \begin{minipage}{0.5\textwidth}
    \centering
    \includegraphics[width=\linewidth]{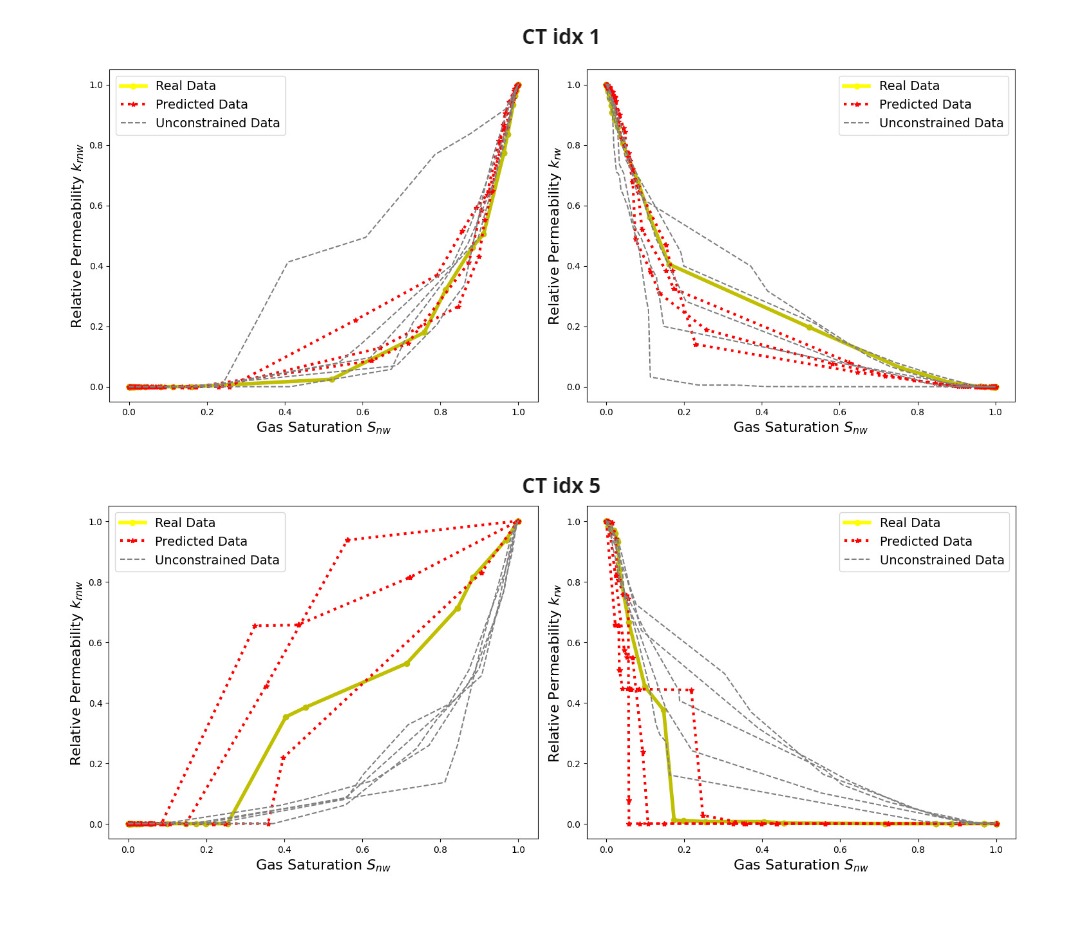}
  \end{minipage}
    \caption{Relative permeability simulations on real porous medium, transformer-assembled porosity-constrained porous medium, and porous medium~(unconstrained by porosity map) at volume dimension $576^3$ corresponding to other CT indices. Most transformer-interpolated $k_r$ curves agree well with ground truth, except for $k_{rw}$ at CT index 0}
  \label{result_fig:kr_ct_0125_vol_9}
\end{figure}

Permeability and relative permeability are less obviously related to porosity distribution. However, our numerical experiments show very promising results that constraining models to spatial porosity distribution can yield fairly accurate prediction of single phase permeability and relative permeability, as indicated in Figure~\ref{result_fig:interpolate_k} - \ref{result_fig:kr_ct_0125_vol_9}. Figure~\ref{result_fig:interpolate_k} shows permeability interpolation at two reconstruction scales: smaller scale $384^3$ and larger scale $576^3$. Smaller scale reconstruction has a total of 30 target samples whereas larger scale is much sparser thus we only have a total of 6 target samples corresponding to 6 CT scans. Both results show fairly accurate predictions of $k_{abs}$ given porosity constraints with an approximately $55$ md mean absolute error. For reference, the ground truth values range from 6 to 600 md. The prediction for sub-volumes within CT index $0$ and $1$ also shows similar accuracy compared with the rest of the training set. Although CT index $0$ and $1$ are not within our training set for both VQVAE and transformer, our porosity-constrained transformer can still successfully predict their values within an acceptable error range.

Unlike absolute permeability, relative permeability is a complex function that captures the complex interplay between fluid phases in porous media. Even for a simple drainage process, relative permeability is a sophisticated saturation-dependent function rather than a single scalar value. It can be conceptualized as a dynamic state function that encapsulates the intricate relationship between fluid phases as they navigate through the pore network. To better evaluate whether spatial porosity constraints do influence the interpolation of drainage relative permeability curves, we not only simulate multiphase flow on target porous medium samples and transformer-generated samples as we did for permeability prediction, but we also obtain $k_r$ data from several sampled porous volumes within CT indices 0-5 to construct uncertainty boundary as we did for the two-point probability function evaluation. Reference data represents typical $k_r$ curves obtained from other core samples, regardless of the constraining porosity value. The expectation is that spatial porosity-constrained transformer-generated porous medium should give a closer $k_r$ prediction within the reference data uncertainty boundary.

Simulation results for two selected volume dimensions ($384^3$ and $576^3$) show promising outcomes. Figure~\ref{result_fig:kr_ct_0_1_vol_6} presents relative permeability simulations for test set CT indices 0 and 1 at $384^3$ volume. The transformer-based ensemble $k_r$ predictions align closely with real $k_r$ curves, constraining them better than the results unconstrained to porosity values, particularly for CT index 1. Similar agreements are observed for training set CT indices ${2,3,4,5}$, as shown in Figures~\ref{result_fig:kr_ct_2_3_vol_6} and \ref{result_fig:kr_ct_4_5_vol_6} in appendix~\ref{appendix:image_reconstruction_simulation}. The only notable deviation is for CT index 0 (Figure~\ref{result_fig:kr_ct_0_1_vol_6}), where both transformer-interpolated $k_{rw}$ and $k_{rnw}$ show less agreement with ground truth, especially $k_{rw}$. This pattern persists in the $576^3$ volume simulations (Figure~\ref{result_fig:kr_ct_0125_vol_9}). However, for CT index 0 at $576^3$, while $k_{rw}$ still deviates from the ground truth, $k_{rnw}$ shows improved agreement compared to the $384^3$ results. For other CT indices at $576^3$, both $k_{rw}$ and $k_{rnw}$ predictions align well with the ground truth. Notably, for CT index 5, where ground truth $k_{rw}$ and $k_{rnw}$ exhibit anomalous behavior deviating significantly from reference data, the transformer-based interpolation successfully captures this unusual pattern.

%%%%%%%%%%%%%%%%%% conclusion %%%%%%%%%%%%%%%%%%%%
\section{Conclusion}
\label{sec:conclusion}
This study presents a novel two-stage deep learning framework for reconstructing 3D porous media at arbitrary pore scales constrained by spatial rock properties, specifically porosity. Our approach combines a Vector Quantized Variational Autoencoder (VQVAE) with a multi-token transformer, enabling the generation of coherent, large-scale porous media representations that honor spatial heterogeneity.

The key methodological contributions of this work are twofold. Firstly, we demonstrate the effectiveness of assembling latent image tokens to create a spatial autoregression model. This approach enables the generation of porous media at scales significantly larger than the training data while maintaining spatial coherence. While transformer-based models have been predominantly used in generating language tokens \cite{2019GPT_2} and compressed video latent frame tokens \cite{2023shiyunyu_videolatent}, our experiments demonstrate the efficacy of using autoregressive transformers to model spatial structure tokens. This finding potentially has numerous applications in 3D material or scene reconstruction. Secondly, our multi-token transformer strategy proves highly effective in representing image patch token sets, offering advantages over traditional single-token transformer generation. This success extends the applicability of multi-token transformer generation to multiple spatial domains that maintain desired spatial relationships.

From a physical transport process modeling perspective, our framework provides valuable insights into the relationship between pore structure parameters and flow functions that regulate the flow of multiphase fluid in porous media, such as relative permeability. Previous research mostly focuses on establishing the qualitative relationship between pore structure and flow functions due to the limited availability and variability of CT scans at the Representative Elementary Volume (REV) scale \cite{2018Meng_krporestructure}. Our results demonstrate a strong correlation between spatial porosity distribution and relative permeability curves. The quantitative relationship, observed through our simulation results, offers a foundation for more detailed theoretical investigations through sensitivity analysis.

Moreover, our work provides a potential solution to the long-standing challenge of relative permeability upscaling. By generating models for 3D porous media at arbitrary scales conditioned to field-scale rock properties, we provide a framework that could potentially bridge the gap between pore-scale physics and reservoir-scale modeling. This approach enables a more rigorous subsurface fluid flow simulation workflow, allowing consistent assignment of field-scale geological and pore-scale transport properties. 

Despite the promising results, our approach has several limitations that are worth discussing. One significant challenge is that the modeling process may generate unrealistic or inconsistent porous media structures, particularly when faced with conditioning inputs that deviate significantly from the training distribution~(very small or large $\phi$). Additionally, the fidelity of the reconstructed porous media is heavily dependent on the quality of image tokens produced by the VQVAE. Imperfections in the VQVAE encoding and decoding process can lead to the loss of fine-scale details or the introduction of artifacts in the final reconstructions. These limitations highlight the importance of careful model training and validation, as well as the need for robust error checking and quality control measures in practical applications.

Future work could explore alternative approaches to address these limitations. One promising direction is the replacement of VQVAE with latent diffusion models for image token reconstruction \cite{2021diffusion_Robin}. This could potentially improve the quality and diversity of generated structures. However, it is important to note that our current VQVAE-based approach offers significant advantages, particularly in scenarios with limited training data and computational resources (using a single NVIDIA A6000 GPU). The discrete codebook of VQVAE provides effective regularization of image features, which is especially valuable when working with smaller datasets typical in geoscience applications, where no existing large digital rock model has been trained to guide the learning process. Another critical consideration for future development is the scalability of this workflow. Our framework has been trained on only four large CT scans with their subvolumes as the training set. This dataset is still considered relatively small for neural networks. To reach production-level performance and reliability, more training data and computational resources will be necessary, and much higher fidelity and more accurate flow function reconstruction is expected. The scaling laws observed in large language models \cite{2020Jared_ScalingLaw} are expected to apply here as well, suggesting that increased data and model capacity will lead to improved performance.

In conclusion, this study demonstrates the potential of deep learning techniques to bridge the gap between pore-scale and field-scale modeling of porous media. By enabling the generation of representative, spatially coherent porous media constrained by field-scale properties, our framework opens new possibilities for more accurate and integrated multiscale modeling in geosciences and reservoir engineering. As computational resources continue to advance, the scalability of this approach promises even greater fidelity and applicability in future research and industry applications.

\section*{Acknowledgments}
This work was partially supported by the Carbon Storage (SMART-CS) Initiative funded by the U.S. Department of Energy’s (DOE) Office of Fossil Energy’s Carbon Storage Research program through the National Energy Technology Laboratory (NETL).

%Bibliography
\newpage
\bibliographystyle{unsrt}  
\bibliography{references}

%%%%%%%%%%%%%% appendix %%%%%%%%%%%%%%%%%%%%
\newpage
\appendix
\renewcommand{\thesection}{\Alph{section}}

\newpage
\section*{Appendix}

\section{VQVAE Architectures and Training Hyperparameters}
\label{appendix:VQVAE_structure_train}

The Vector Quantised Variational Autoencoder (VQVAE) architecture and training hyperparameters are presented in Table~\ref{tab:VQVAE_pars}. As illustrated in Figure~\ref{method_fig:VQVAE_workflow}, the VQVAE architecture consists of three main components: an encoder, a codebook, and a decoder. The encoder transforms the input image into a latent representation through a series of 3D downsampled convolutional residual blocks. Each residual block comprises two 3D convolutional layers with a skip connection, incorporating group normalization and Swish activation functions. This structure facilitates gradient flow and enables the training of deeper networks by mitigating the vanishing gradient problem. The codebook then quantizes the encoder's output to a discrete set of learned embedding indices, with a codebook size of 3000. Finally, the decoder, mirroring the encoder's structure with residual blocks and transposed convolutions, reconstructs the image from the quantized latent vectors $z_q$. 

The primary challenge in training a VQVAE model lies in successfully developing a codebook that embodies all effective feature basis latent vectors to characterize segmented subvolumes of porous media while ensuring high reconstruction quality. In other words, the challenge is in balancing the different loss terms in Equation~\ref{equ:lossvqvae}. To stabilize training, we gradually increase the training weights applied to the codebook loss during VQVAE training. We start with an initial codebook loss weight of 0.02 and increase it by 0.02 per epoch, as indicated by the codebook weight increase hyperparameter. This approach prioritizes training the encoder and decoder initially to minimize reconstruction loss, while gradually encouraging the codebook and encoder-decoder representations to converge. Although we set the maximum weight for training codebook is $2$, we found the optimal epochs for VQVAE to have reasonable reconstruction is $25$, where during the last epoch, codebook weight is $0.5$. If we continue training while increasing codebook weight, the reconstruction performance will decrease.

\begin{table}[h] 
\centering
\caption{VQVAE Hyperparameters}
\label{tab:VQVAE_pars}
\begin{tabular}{lll}
\hline
\textbf{Component} & \textbf{Parameter} & \textbf{Value} \\
\hline
\multirow{5}{*}{Encoder} & Image Channels & 1 \\
 & Latent Dimension & 256 \\
 & Number of Residual Blocks & 2 \\
 & Number of Groups & 16 \\
 & Channels & [16, 64, 128, 256, 512] \\
\hline
\multirow{3}{*}{Codebook} & Size & 3000 \\
 & Latent Dimension & 256 \\
 & Beta (Commitment Loss Weight) & 1 \\
\hline
\multirow{5}{*}{Decoder} & Image Channels & 1 \\
 & Latent Dimension & 256 \\
 & Number of Residual Blocks & 3 \\
 & Number of Groups & 16 \\
 & Channels & [512, 256, 256, 64, 16, 16] \\
\hline
\multirow{8}{*}{Training} & Epochs & 25 \\
 & Batch Size & 20 \\
 & Learning Rate (VQGAN) & 0.0005 \\
 & Beta1 & 0.9 \\
 & Beta2 & 0.999 \\
 & Initial Codebook Loss Weight & 0.02 \\
 & Codebook Weight Increase (per epoch) & 0.02 \\
 & Maximum Codebook Loss Weight & 2 \\
\hline
\end{tabular}
\end{table}

\newpage
\section{Transformer Architecture and Training Hyperparameters}
\label{appendix:transformer_structure_training}

The transformer model in our work is primarily based on the NanoGPT~(reproduction of GPT-2 \cite{2019GPT_2}) architecture developed by Andrej Karpathy, with modifications to suit our specific task. As detailed in Table~\ref{tab:transformer_params}, the model consists of 12 layers, each with 12 attention heads, and an embedding dimension of 1080. The architecture employs a vocabulary size of 3001, which corresponds to the VQVAE's codebook size plus an additional token for the start-of-sequence. The model employs an attention window size of 512, which corresponds to 8 token sets $\mathcal{S}$, where each set contains 64 image tokens that characterize an individual porous medium subvolume with size $64^3$ voxels. This configuration allows the model to process and correlate information from 8 spatial adjacent porous medium subvolumes in a 2 cubic spatial setting, meaning the full attention window encompasses a total volume of $128^3$ voxels ($2 \times 2 \times 2$ subvolumes). We only add porosity as our conditional information for each subvolume, with a conditional dimension of 1 and a conditional embedding size of 100. The architecture includes several custom components such as CausalSelfAttention, MLP, and Block modules, which work together to process the input tokens and generate output logits.

\begin{table}[h]
\centering
\caption{Transformer Architecture and Training Hyperparameters}
\label{tab:transformer_params}
\begin{tabular}{lll}
\hline
\textbf{Component} & \textbf{Parameter} & \textbf{Value} \\
\hline
\multirow{11}{*}{Architecture} & Block Size & 512 \\
 & Vocabulary Size & 3001 \\
 & Number of Layers & 12 \\
 & Number of Attention Heads & 12 \\
 & Embedding Dimension & 1080 \\
 & Dropout Rate & 0.01 \\
 & Use Bias & True \\
 & Conditional Dimension & 1 \\
 & Conditional Embedding Size & 100 \\
 & Token Embedding Size & 3000 \\
 & Number of Features & 64 \\
\hline
\multirow{4}{*}{Training} & SOS token & 3000 \\
 & Batch Size & 32 \\
 & Number of Epochs & 50 \\
 & Learning Rate & 0.0002 \\
\hline
\end{tabular}
\end{table}

\newpage
\section{Supplementary Image Reconstructions and $k_r$ Simulation Results}
\label{appendix:image_reconstruction_simulation}

More image reconstructions and $k_r$ simulation results have been appended in this section for further inspection of our workflow effectiveness.

\begin{figure}[hbt]
    \centering
    \includegraphics[width = \textwidth]{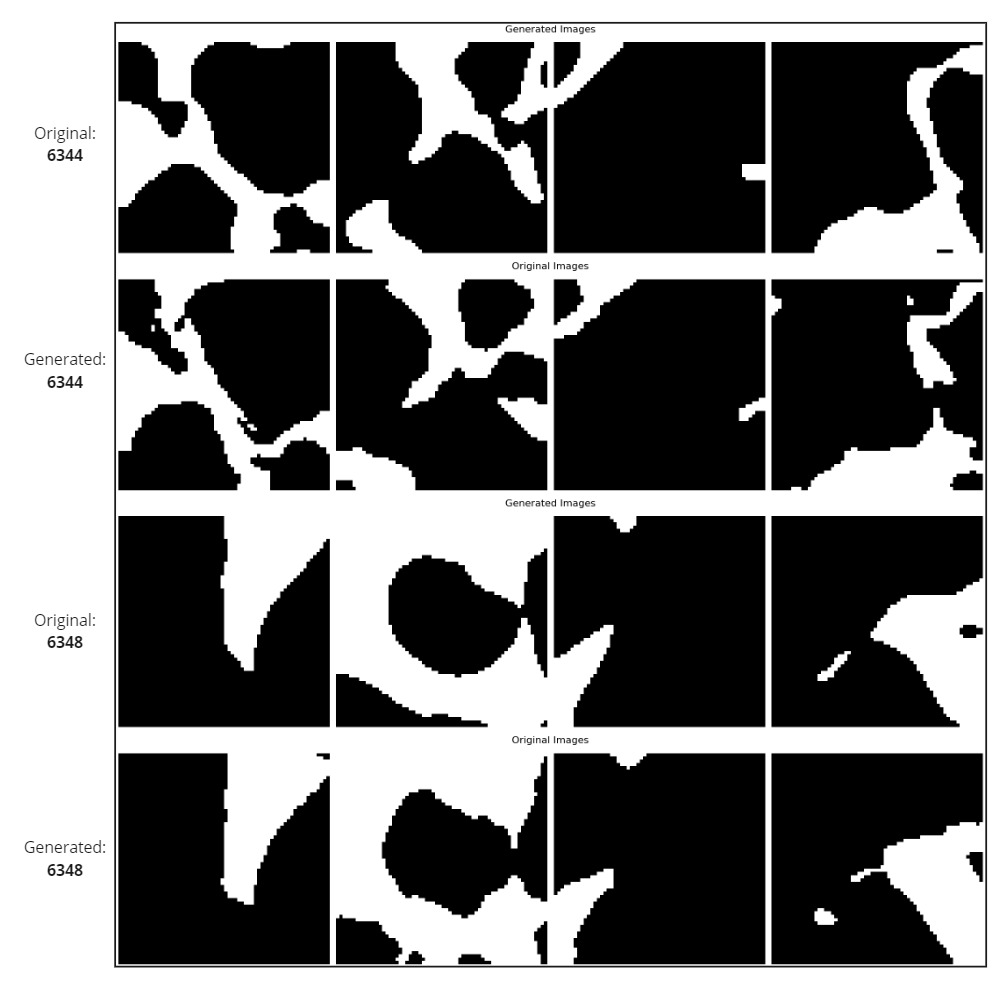}
    \caption{Original segmented image versus Decoded images by VQVAE; The first two rows represent the comparison among image patches on CT index 0. The last two rows represent the comparison among image patches on CT index 1. There is a small discrepancy in sample $6344$ where the most right synthetic pore space loses the connection whereas the original structure keeps such connection. This discrepancy may cause fluctuations of $k_{abs}$ due to sparse percolation paths but is less worrisome for the further assembled porous mediums which have a much larger simulation domain.
}
    \label{method_fig:VQVAE_reconstruction_compare}
\end{figure}\par

\begin{figure}[hbt]
    \centering
    \includegraphics[width = 0.85\textwidth]{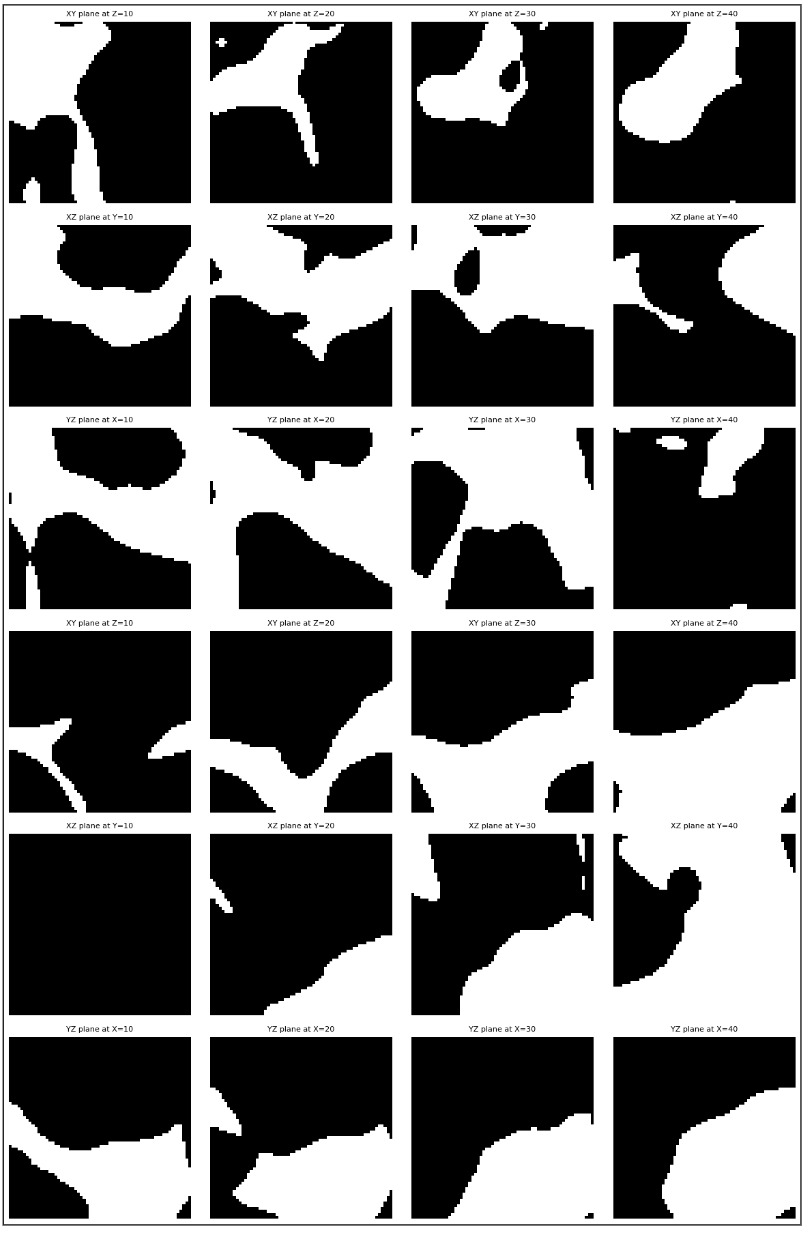}
    \caption{Transformer sampled image patches. The first three rows show the visualization of one patch's different sections - XY, XZ, YZ planes. The last three rows show the visualization of another image patch of different section visualizations
}
\label{result_fig:transformer_64_img}
\end{figure}\par

\begin{figure}[hbt]
  \centering
  \begin{minipage}{0.4\textwidth}
    \centering
    \includegraphics[width=\linewidth]{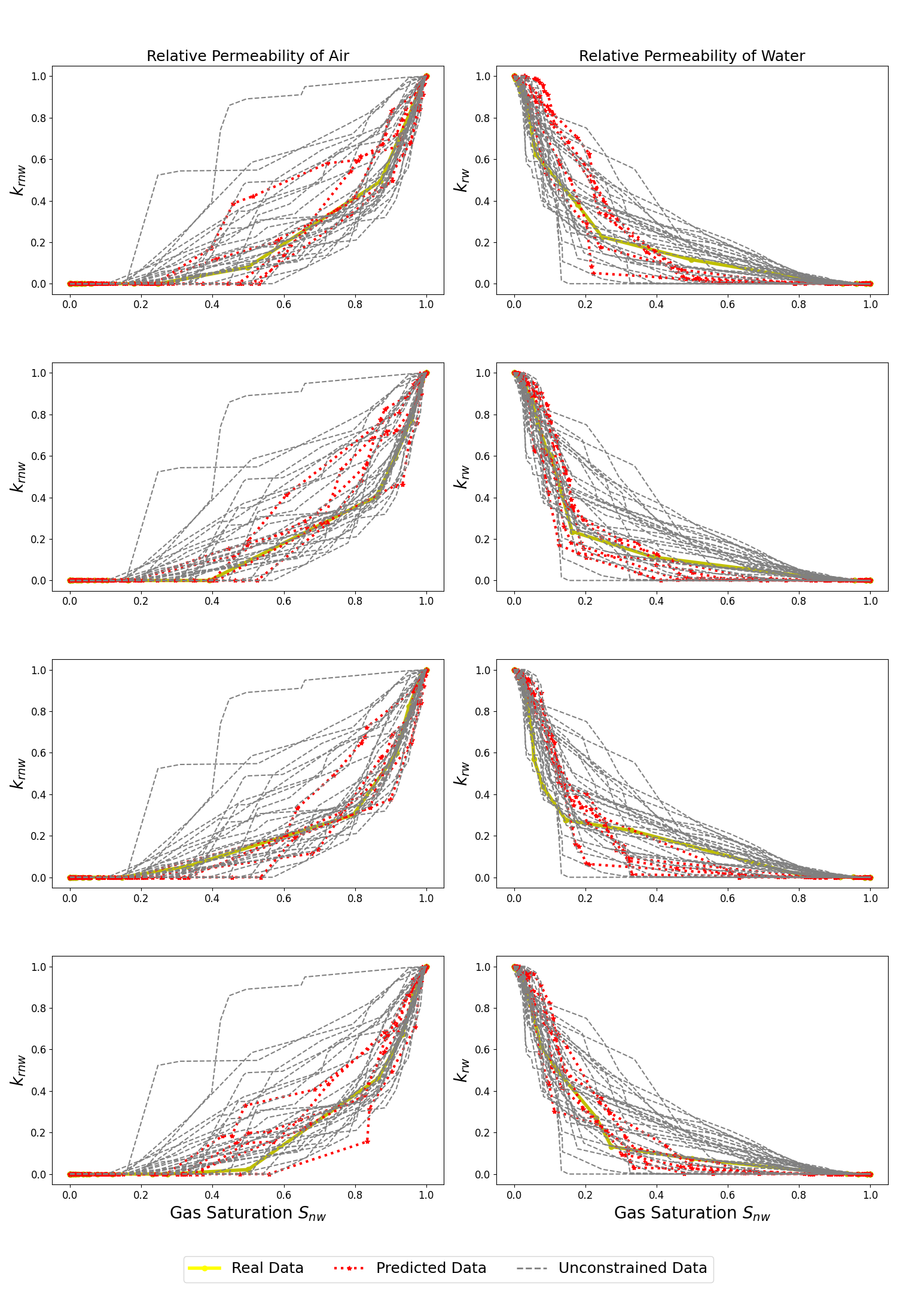}
  \end{minipage}%
  \hfill
  \begin{minipage}{0.4\textwidth}
    \centering
    \includegraphics[width=\linewidth]{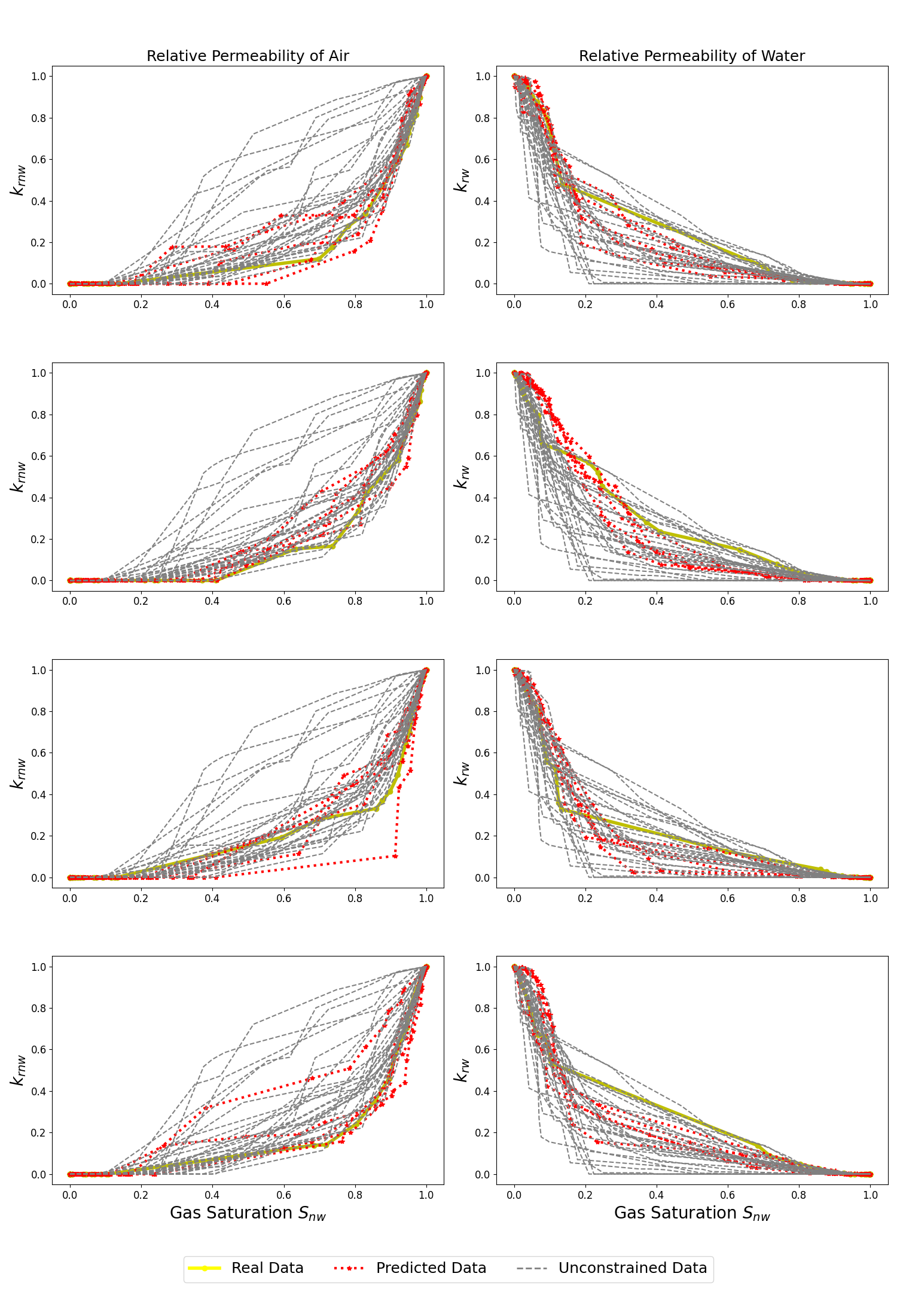}
  \end{minipage}
  \caption{Relative permeability simulations on real porous medium, transformer-assembled porosity-constrained porous medium, and reference porous medium at volume dimension $384^3$. Left: $k_r$ simulation on CT index 2; Right: $k_r$ simulation on CT index 3. Transformer-interpolated $k_r$ agrees well with ground truth curves on both CT index 2 and 3}
  \label{result_fig:kr_ct_2_3_vol_6}
\end{figure}

\begin{figure}[hbt]
  \centering
  \begin{minipage}{0.4\textwidth}
    \centering
    \includegraphics[width=\linewidth]{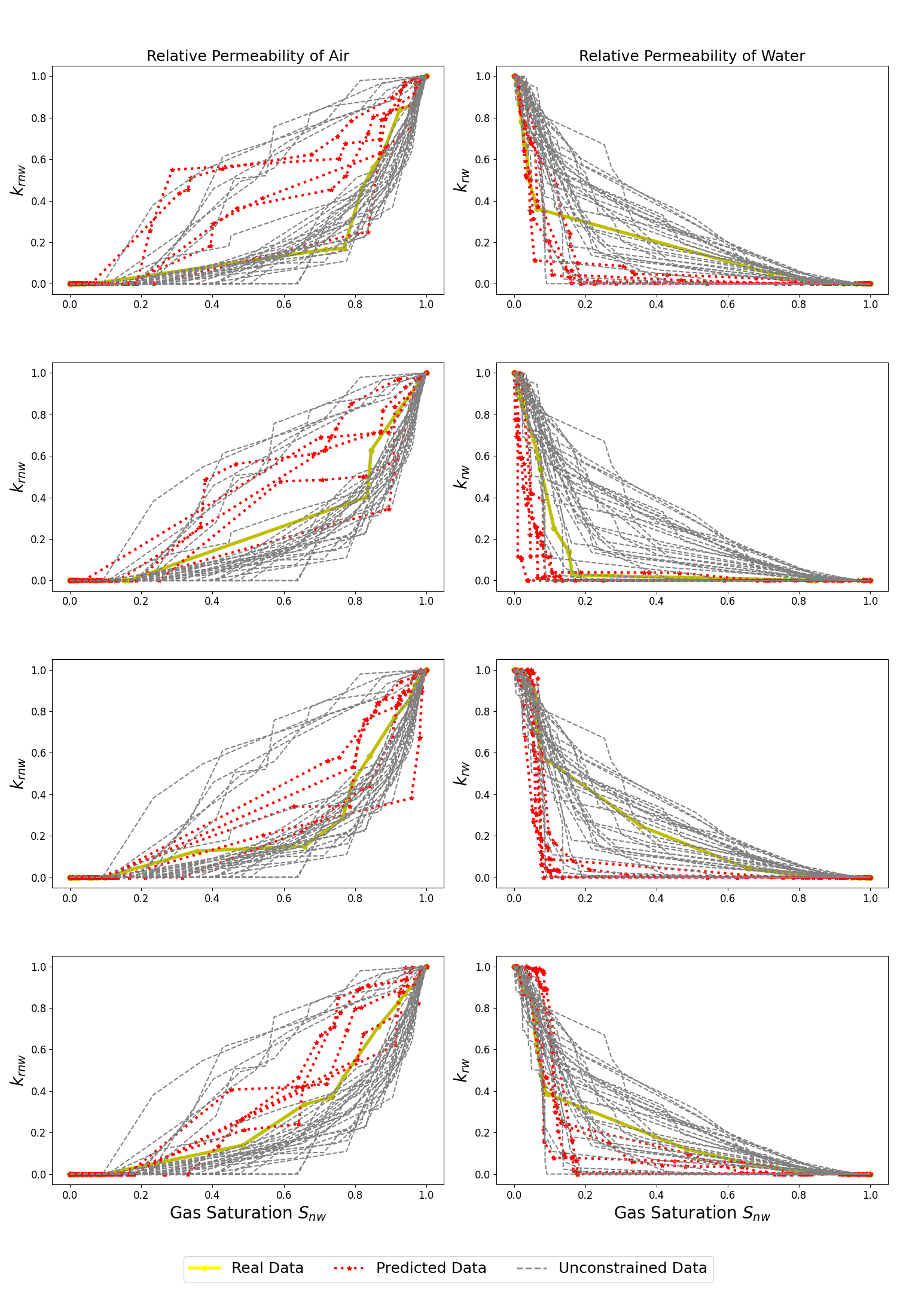}
  \end{minipage}%
  \hfill
  \begin{minipage}{0.4\textwidth}
    \centering
    \includegraphics[width=\linewidth]{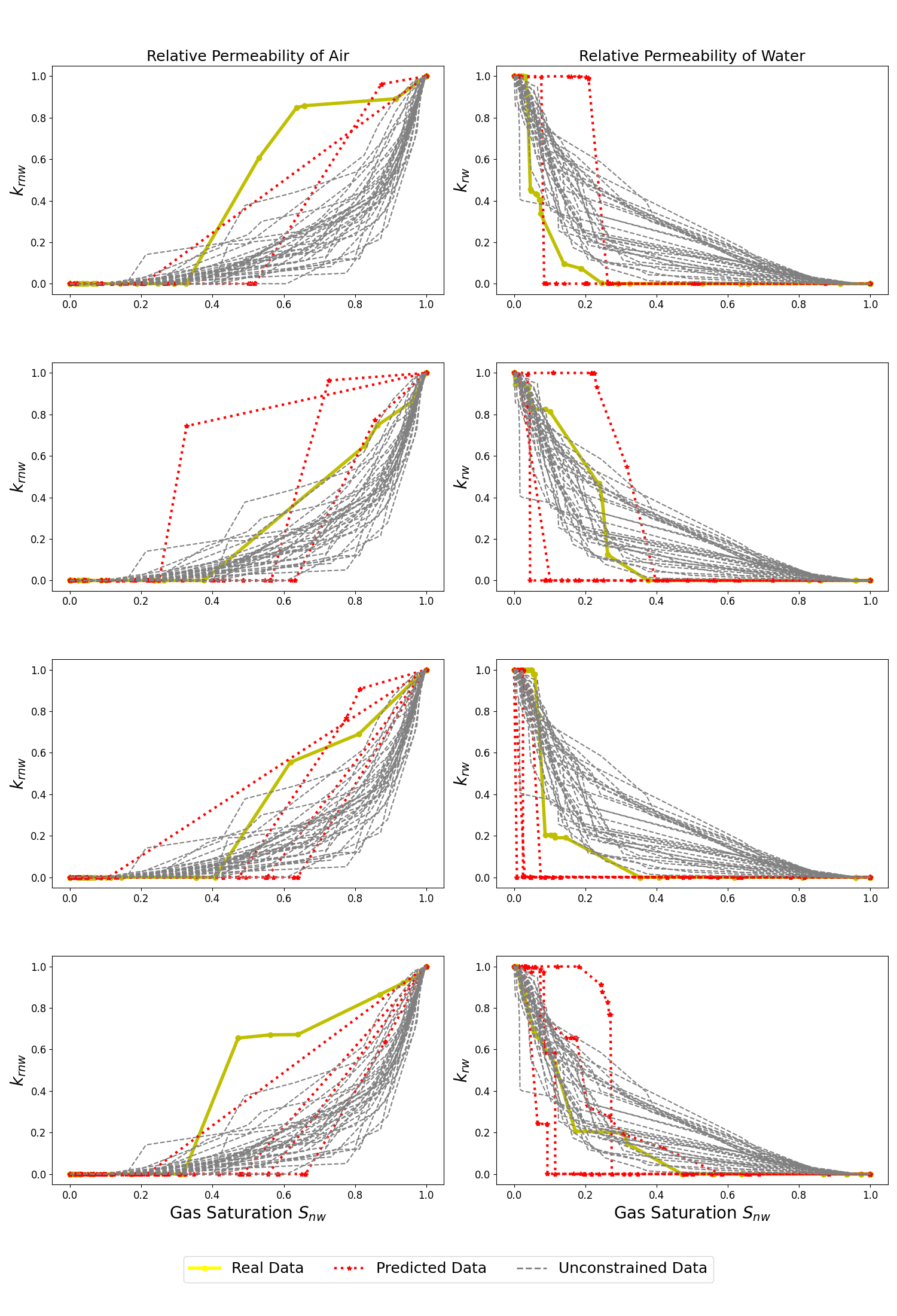}
  \end{minipage}
    \caption{Relative permeability simulations on real porous medium, transformer-assembled porosity-constrained porous medium, and reference porous medium at volume dimension $384^3$. Left: $k_r$ simulation on CT index 4; Right: $k_r$ simulation on CT index 5. Transformer-interpolated $k_r$ agrees well with ground truth curves for most ensemble realizations at both CT indices 4 and 5; However, there is some disagreement for $k_{rw}$ at CT index 4}
    \label{result_fig:kr_ct_4_5_vol_6}
\end{figure}

\end{document}